\documentclass[10pt,twocolumn,letterpaper]{article}

\usepackage{template}              %
\usepackage{pifont}

\usepackage{multirow}
\usepackage{pifont}
\usepackage{multirow}
\usepackage{graphicx}
\usepackage{subcaption}
\usepackage{xcolor}
\usepackage{colortbl}
\usepackage{fixmetodonotes}
\usepackage{bm}

\newcommand{\cmark}{\ding{51}} %
\newcommand{\xmark}{\ding{55}} %
\newcommand{\method}{NAF}

\newif\ifshowedits

\newcommand{\addeditor}[3]{%
  \definecolor{#1color}{rgb}{#3}
  \expandafter\newcommand\csname #1\endcsname[1]{%
  \ifshowedits
    {\color{#1color} ##1}%
  \else
    {##1}%
  \fi
  }%
  \expandafter\newcommand\csname #1rmk\endcsname[1]{%
  \ifshowedits
    {\color{#1color} {\bf [#2: ##1]}}
  \fi
  }%
  \expandafter\newcommand\csname #1rpl\endcsname[2]{%
  \ifshowedits
    {\color{#1color} ##1 \sout{##2}}
  \else
    {##1}
  \fi
  }%
}

\newcommand{\best}[1]{\textbf{#1}}
\newcommand{\sbest}[1]{\underline{#1}}
\newcommand{\gain}[1]{\textcolor{ForestGreen}{#1}}
\newcommand{\bestgain}[1]{\textcolor{ForestGreen}{\textbf{#1}}}
\newcommand{\sbestgain}[1]{\textcolor{ForestGreen}{\underline{#1}}}
\newcommand{\loss}[1]{\textcolor{BrickRed}{\textbf{#1}}}
\showeditstrue
\newcommand{\oomcell}{\cellcolor{red!10}\textcolor{BrickRed}{\textbf{OOM}}}

\usepackage{cuted} %
\usepackage{titletoc} %

\definecolor{bblue}{rgb}{0.21,0.49,0.74}

\definecolor{darkpastelblue}{rgb}{0.47, 0.62, 0.8}
\definecolor{lightpastelblue}{rgb}{0.75, 0.85, 0.95}
\usepackage[pagebackref,breaklinks,colorlinks,allcolors=bblue]{hyperref}

\title{NAF: Zero-Shot Feature Upsampling via Neighborhood Attention Filtering}

\author{%
Loïck Chambon\textsuperscript{1,2} 
\hspace{2.5mm} 
Paul Couairon\textsuperscript{2}
\hspace{2.5mm}
\'Eloi Zablocki\textsuperscript{1}
\\
\hspace{2.5mm} 
Alexandre Boulch\textsuperscript{1}
\hspace{2.5mm} 
Nicolas Thome\textsuperscript{2,3}
\hspace{2.5mm} 
Matthieu Cord \textsuperscript{1,2}
\vspace{2.5mm} 
\and
\textsuperscript{1}Valeo.ai, Paris, France \hspace{2.5mm} 
\textsuperscript{2}Sorbonne Universit\'e, CNRS, ISIR, F-75005 Paris, France \\
\textsuperscript{3}Institut Universitaire de France (IUF)
\vspace{-5cm}
}

\begin{document}

\maketitle
\begin{strip}
\vspace{-0.2cm}
\centering
    \includegraphics[width=0.9\textwidth]{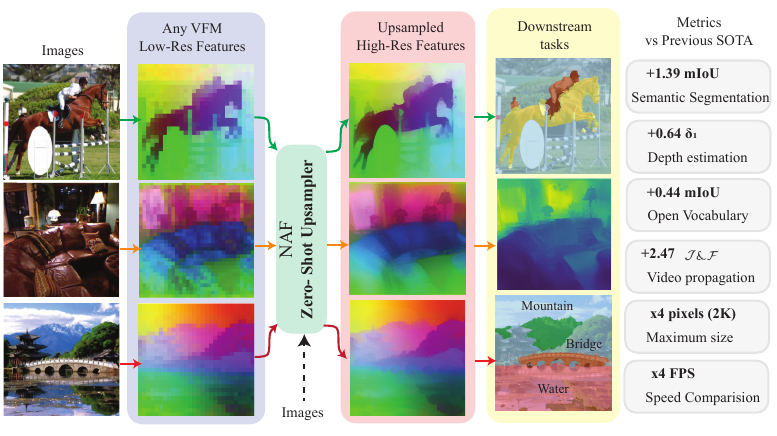}
\captionof{figure}{\textbf{Neighborhood Attention Filtering (NAF) as a Zero-Shot Feature Upsampler}: train once, apply efficiently to any Vision Foundation Model (including 7B models) to any scale, achieving state-of-the-art results across multiple downstream tasks.}
\label{fig:teasing}
\end{strip}

\begin{abstract}
Vision Foundation Models (VFMs) extract spatially downsampled representations, posing challenges for pixel-level tasks.
Existing upsampling approaches face a fundamental trade-off: classical filters are fast and broadly applicable but rely on fixed forms, while modern upsamplers achieve superior accuracy through learnable, VFM-specific forms at the cost of retraining for each VFM.
We introduce Neighborhood Attention Filtering (NAF), which bridges this gap by learning adaptive spatial-and-content weights through Cross-Scale Neighborhood Attention and Rotary Position Embeddings (RoPE), guided solely by the high-resolution input image.
NAF operates zero-shot: it upsamples features from any VFM without retraining, making it the first VFM-agnostic architecture to outperform VFM-specific upsamplers and achieve state-of-the-art performance across multiple downstream tasks.
It maintains high efficiency, scaling to 2K feature maps and reconstructing intermediate-resolution maps at 18 FPS.
Beyond feature upsampling, NAF demonstrates strong performance on image restoration, highlighting its versatility. Code and checkpoints are available at \url{https://github.com/valeoai/NAF}.
\end{abstract}

\section{Introduction}
\label{sec:intro}

Vision Foundation Models (VFMs) extract rich semantic representations from images.
However, these features are generated at reduced spatial resolutions due to computational constraints and architectural choices, limiting their effectiveness for fine-grained tasks.
Increasing input image size is a straightforward approach, but only few VFMs exhibit scale invariance and handle non-standard resolutions effectively \citep{heinrich2025radiov25,ranzinger2023amradio,simeoni2025dinov3,touvron2020fixefficientnet}. For most VFMs, this degrades performance \citep{jafar}, and computational cost scales quadratically with resolution, causing prohibitive inference times and memory constraints.
A more promising approach directly upsamples VFM output features rather than enlarging input images \citep{kopf2007jbu,lift,featup,featsharp,jafar,wimmer2025anyup}.

Early approaches employ filtering techniques to upsample features, relying on spatial proximity \citep{keys2003bicubic,duchon1979lanczos,parker2007comparison} or incorporating guidance from the input image \cite{tomasi1998bilateralfilter,he2012guidedfilter,he2015fastguidedfilter,kopf2007jointbilateral}.
However, traditional filters' reliance on fixed forms (e.g., Gaussian) limits their adaptability, often leading to suboptimal results.
To overcome this, learning-based feature upsamplers \citep{featup,lift,loftup,featsharp,jafar,wimmer2025anyup} have been introduced. They optimize their parameters to recover high-resolution features from low-resolution ones.
While achieving higher-quality upsampling, they lack of interpretability and sacrifice efficiency of classical filters, introducing complex pipelines with low throughput, high memory consumption, and large parameter counts (see. \autoref{tab:caracteristics}), resulting in relatively modest maximum upscaling, or worse, fixed ratios \citep{featsharp, featup, lift}.
{Crucially, existing upsamplers depend on VFM-specific guidance, which forces retraining whenever the underlying VFM changes.}

We introduce Neighborhood Attention Filtering (\method{}), a VFM-agnostic upsampling module that generalizes in a zero-shot manner to features from any VFM.
\method{} reweights features using image-based guidance and relative spatial proximity (see \autoref{fig:teasing}) via Cross-Scale Neighborhood Attention for local feature similarity, combined with Rotary Position Embeddings (RoPE) to encode relative spatial relationships.
{We show that this design implicitly learns an Inverse Discrete Fourier Transform (IDFT) of the aggregation: \method{} predicts spectral coefficients that reconstruct an adaptive, data-dependent upsampling filter.
This formulation preserves the interpretability of classical filters while allowing the model to learn flexible, spatial-and-content-aware aggregation (see. \autoref{sec:supp_maths}).}

Overall, our contributions are:
\begin{itemize}
\item We present $\text{\method{}}$, a VFM-agnostic feature upsampler guided solely by high-resolution images. It leverages a Cross-Scale Neighborhood Attention mechanism for content-aware local interpolation, exhibiting strong similarities with traditional filtering methods and IDFT.

\item We show that \method{} achieves state-of-the-art performance across diverse vision tasks and datasets, for many VFM families and a wide range of model sizes.

\item We implement an efficient upsampler that performs zero-shot upsampling at high throughput, to high resolutions (up to 2K) and works on very large VFMs (7B parameters), which previous methods cannot handle, while still achieving notable gains.

\item We demonstrate \method{}'s generalization beyond upsampling to tasks such as image denoising, using the same architecture and opening new avenues for cross-task feature filtering.

\end{itemize}

\section{Background and Related Work}
\label{sec:related_work}

\paragraph{Background: Filtering.} Filtering is a fundamental operation in computer vision for recovering spatial details. In feature upsampling, we consider a pair of low-resolution (LR) and high-resolution (HR) feature maps ($\mathbf{F}^{\mathrm{LR}}$, $\mathbf{F}^{\mathrm{HR}}$), both of dimension $d\in\mathbb{N}$. The objective is to reconstruct, for each location $p$ of the high-resolution map, the corresponding feature representation $\mathbf{F}^{\mathrm{HR}}_{p}$. 

\smallskip\noindent\emph{Spatial-and-content-aware filters.} While spatial-aware filters~\citep{keys2003bicubic,duchon1979lanczos,unser1991fastcubicbspline,ruijters2012gpucubicbspline,zhang2014medianfilter} reweighting the input based on spatial proximity remain widely used, spatial-and-content-aware filters improve upon them by leveraging an auxiliary guidance signal $\mathbf{G}$ (e.g., an image) to preserve edges and fine structures. The high-resolution feature at location $p$ is computed as
\begin{equation}
\mathbf{F}_{p}^{\mathrm{HR}} = \frac{1}{Z(p)} \sum_{q \in \mathcal{N}(p)} w(p, q \,|\, \mathbf{G}) \mathbf{F}^{\mathrm{LR}}_{q},
\label{eq:filter}
\end{equation}
where $\mathcal{N}(p)$ is a local neighborhood, $Z(p)$ is a normalization factor, and $w(p, q \,|\, \mathbf{G})$ are content-adaptive weights reflecting similarity in the guidance signal. Classic examples include the bilateral filter~\citep{tomasi1998bilateralfilter}, joint bilateral filter~\citep{kopf2007jointbilateral} (JBF) and others~\citep{he2012guidedfilter,he2015fastguidedfilter}, where weights depend on both spatial proximity and intensity similarity: $w(p, q \,|\, \mathbf{G}) = \exp\Big( - \frac{\|p - q\|^2}{2\sigma_s^2} - \frac{\|\mathbf{G}_{p} - \mathbf{G}_{q}\|^2}{2\sigma_r^2} \Big)$, with $\sigma_s$ and $\sigma_r$ controlling spatial and range intensity respectively. While effective at preserving edges, traditional formulations are limited by fixed kernel shapes and handcrafted similarity functions, motivating learning-based adaptive filters that learn expressive, content-dependent weights directly from data.

\begin{table}[t]
\centering
\resizebox{\columnwidth}{!}{%
\begin{tabular}{@{}lcc cc c@{}}
\toprule
\multirow{2}{*}{\textbf{Method}} & \textbf{VFM} & \textbf{\# Params} & \multirow{1}{*}{\textbf{GFLOPs}} & \multirow{1}{*}{\textbf{FPS}} & \textbf{Maximum} \\
\cmidrule(lr){4-5}
& \textbf{Agnostic} & \textbf{(Million)} & \multicolumn{2}{c}{\emph{computed at $\times 16$}} &  \textbf{Ratio} \\
\midrule
FeatUp  \cite{featup}   & \xmark & 0.17 & 83 & 17 & 16 \\
LiFT \cite{lift}    & \xmark & 1.19 & 512 & 30 & 16 \\
JBU \citep{kopf2007jbu} & \cmark & 0.03 & 4.88 & 4 & 28 \\
LoftUp  \cite{loftup}   & \xmark & 4.30  & 1971 & 4 & 32 \\
AnyUp \citep{wimmer2025anyup} & \cmark & 0.88 & 329   & 5 & 32 \\
JAFAR   \cite{jafar}   & \xmark & 0.63    & 366  & 11 & 32 \\
\rowcolor{lightpastelblue} \textbf{\method{}}  & \cmark & 0.66 & 265  & 18 & 72 \\
\bottomrule
\end{tabular}
}
\caption{\textbf{Comparison of upsampling methods} for a $\times16$ upsampling of input features $(384,28,28)$.  
The maximum ratio indicates the largest upscaling factor that fits on a single A100 40GB GPU.  
Methods are sorted by maximum ratio and FPS.}
\label{tab:caracteristics}
\end{table}

\paragraph{Feature Upsampling.}
Deep learning naturally extends classical filtering by allowing the aggregation operation to be learned end-to-end through parameterized kernels. These deep methods optimize the combination of low-resolution features $\mathbf{F}^{\mathrm{LR}}$ using guidance $\mathbf{G}$, which typically includes an encoding of the input image $\mathbf{I}$ and the low-resolution features themselves. This leads to the generic formulation:
\begin{equation}
\mathbf{F}^{\mathrm{HR}}_{p}
= \frac{1}{Z(p)} \sum_{q \in \mathcal{N}(p)} 
w_{\theta'}(p, q \,|\, \operatorname{Enc}_\theta (\mathbf{I}), \mathbf{F}^{\mathrm{LR}}_{q}) \mathbf{F}^{\mathrm{LR}}_{q},
\end{equation}
with $\theta'$ learnable parameters of the kernel and $\operatorname{Enc}_\theta$ a trainable image encoder.
The development of feature upsampling methods is largely driven by the specific context in which they are deployed, particularly whether they are designed for specific tasks or general vision backbones.

\smallskip\noindent\emph{Task-Designed Upsamplers.}
These methods serve as integrated components within downstream deep learning pipelines, such as semantic segmentation and depth estimation, which require pixel-level precision. Early techniques rely on standard methods like transposed convolutions or pixel unshuffling \citep{li2018pyramid,ronneberger2015unet,shi2016pixelunshuffling,zhao2017pspnet,long2015fcn}. More sophisticated approaches use adaptive reweighting: CARAFE \citep{carafe} and DySample \citep{dysample} predict content-aware kernels or sampling points, while SAPA \citep{sapa} exploits local similarity between high-resolution guidance and low-resolution features. While effective, these task-specific upsamplers require retraining for every new downstream task and are not primarily designed for general VFM features.

\smallskip\noindent\emph{VFM-Specific Upsamplers.}
These upsamplers are explicitly aligned with the feature distribution of a particular VFM and are trained to generalize across different vision tasks without task-specific supervision. FeatUp \citep{featup} and LiFT \citep{lift} introduced early dedicated pipelines: FeatUp relies on a parameterized variant of Joint Bilateral Upsampling (JBU) module \citep{kopf2007jbu}, and LiFT uses a CNN-based encoder-decoder architecture. FeatSharp \citep{featsharp} builds on JBU by adding tiling and debiasing strategies. More recently, JAFAR \citep{jafar} and LoftUp \citep{loftup} introduced attention-based architectures which allows for continuous upsampling to arbitrary resolutions. Crucially, all these methods rely heavily on the semantic content of the VFM's low-resolution features to compute the upsampling guidance.

\smallskip\noindent\emph{VFM-Agnostic Upsamplers.}
The most challenging goal is to create upsamplers that can be applied in a zero-shot manner to features from \emph{any} VFM without the need for retraining. The Joint Bilateral Upsampling (JBU) module \citep{featup} is inherently VFM-agnostic, but its performance as a standalone upsampler is limited compared to modern methods.
A recent concurrent work, AnyUp \citep{wimmer2025anyup}, builds on attention mechanisms \citep{loftup, jafar} and removes the dependency on feature dimensionality allowing to be used to multiple VFMs. However, AnyUp's guidance computation still relies on the VFM's low-resolution features, meaning the upsampling remains a priori entangled with the target feature distribution. Furthermore, it is computationally heavier than the current best VFM-specific upsamplers \autoref{tab:caracteristics}. %
In contrast, \method{} achieves full VFM-agnosticism by deriving guidance solely from a lightweight encoder applied to the input image $\mathbf{I}$, eliminating any dependency on VFM low-resolution features. This simple yet effective design makes \method{} roughly $\times 4$ faster than AnyUp with $25\%$ fewer parameters, while achieving superior reconstruction quality.

\section{Method: NAF}
\subsection{Architecture}
\label{sec:method:naf}

We introduce Neighborhood Attention Filtering (NAF), a learnable upsampling framework that generalizes classical filtering through an attention formulation.
NAF interprets upsampling as a spatial- and content-aware aggregation of low-resolution features, guided solely by the input image.
{Instead of predicting the spatial aggregation kernel directly, NAF learns its representation in the frequency domain. In practice, it predicts the spectral coefficients whose inverse discrete Fourier transform gives the spatial kernel, as we show in \autoref{sec:supp_maths}.}

\paragraph{General formulation.}

Starting from an input image $\mathbf{I} \in \mathbb{R}^{H_{\mathrm{HR}} \times W_{\mathrm{HR}} \times 3}$, a vision foundation model (VFM) produces a low-resolution feature map $\mathbf{F}^{\mathrm{LR}} \in \mathbb{R}^{H_{\mathrm{LR}} \times W_{\mathrm{LR}} \times d}$. The goal of NAF is to reconstruct a high-resolution feature map $\mathbf{F}^{\mathrm{HR}} \in \mathbb{R}^{H_{\mathrm{HR}} \times W_{\mathrm{HR}} \times d}$ that aligns with the fine spatial details of the image $\mathbf{I}$, where $H_{\mathrm{HR}} = s \cdot H_{\mathrm{LR}}$ and $W_{\mathrm{HR}} = s \cdot W_{\mathrm{LR}}$ for an upsampling factor $s$.

At its core, NAF computes the high-resolution feature at position $p$ as an attention-weighted combination of low-resolution features in its spatial neighborhood $\mathcal{N}(p)$:
\begin{equation}
\mathbf{F}^{\mathrm{HR}}_{p} =
\frac{1}{Z(p)}
\sum_{q \in \mathcal{N}(p)}
\operatorname{exp}\left(\frac{\langle Q_p, K_q \rangle}{\sqrt{d}} \right) \mathbf{F}^\mathrm{LR}_{q},
\label{eq:naf}
\end{equation}

\begin{figure}
    \centering
    \includegraphics[width=\columnwidth]{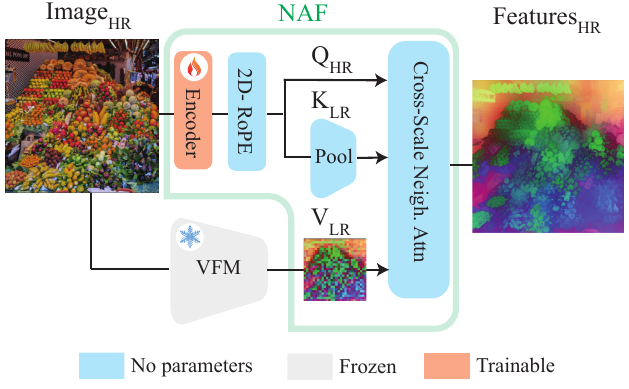}
    \caption{\textbf{\method{} architecture} allows to upsample low-resolution VFM features to any resolution, guided solely by the original high-resolution image.}
    \label{fig:architecture}
\end{figure}

where $Q$, $K$ denote queries and keys, $\langle. , .\rangle$ defines the dot product, and $Z(p)$ is a normalization factor. The attention \emph{values} correspond directly to the VFM features $F^\text{LR}$.
The key design question is therefore: \emph{how to define $Q$, $K$ so that the attention captures cross-scale similarity while remaining independent to the VFM?}

The overall NAF architecture is illustrated in \autoref{fig:architecture} and detailed in the following.

\paragraph{Dual-Branch Guidance Encoder.}
Queries and keys are both derived from the input image $\mathbf{I}$ through a learnable Dual-Branch Guidance Encoder that extracts a high-resolution guidance map:
$\text{Enc}_{\theta}(\mathbf{I}) \in \mathbb{R}^{H_{\mathrm{HR}} \times W_{\mathrm{HR}} \times C}$,
where $\theta$ is a set of learned parameters and $C$ denotes the number of guidance channels.

\begin{figure}
    \centering
    \includegraphics[width=\columnwidth]{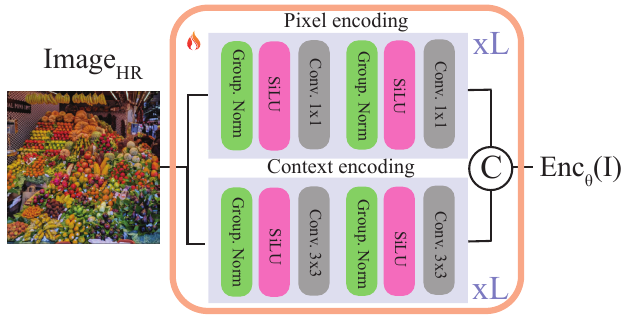}
    \caption{\textbf{Details of the dual-branch image encoder.} \method{} encoder considers both a pixel-wise branch and a local-contextual branch.}
    \label{fig:encoder}
\end{figure}

{Inspired by Inception design~\cite{szegedy2015inception}, the}
encoder is composed of two complementary branches designed to capture both fine-grained pixel details and local contextual information (see \autoref{fig:encoder}). The pixel-encoding branch applies a series of $1{\times}1$ convolutional blocks to extract pixel-wise features, while the contextual-encoding branch employs $3{\times}3$ convolutions to aggregate local neighborhood information.
Each branch consists of $L$ stacked convolutional blocks and outputs $C/2$ channels. The resulting feature maps from both branches are concatenated along the channel dimension.

\paragraph{RoPE and Pooling.}
To encode relative positional information, we apply 2D Rotary Positional Embeddings (RoPE) \citep{heo2024rotary} to the guidance features, yielding position-aware features $\operatorname{RoPE}(\operatorname{Enc}_\theta(\mathbf{I}))$.

Attention \emph{queries} $Q$ correspond to the high-resolution RoPE-encoded features:
\begin{equation}
Q_{p} := \operatorname{RoPE}(\text{Enc}_\theta(\mathbf{I}))_p.
\label{eq:queries}
\end{equation}

Attention \emph{keys} $K$ are obtained by average pooling the same features to the low-resolution grid, ensuring geometric alignment with the low-resolution features $\mathbf{F}^\text{LR}$:
\begin{equation}
K_{q} := \operatorname*{AvgPool}_{q' \in q}\big[ \operatorname{RoPE}(\text{Enc}_\theta(\mathbf{I}))_{q'} \big],
\label{eq:keys}
\end{equation}
where the pooling is taken over all high-resolution pixel $q'$ falling within the low-resolution position $q$.

\paragraph{Cross-Scale Neighborhood Attention.}
Dense visual features exhibit strong spatial autocorrelation, with the most informative cues for upsampling a pixel lying within its local vicinity. Prior work~\cite{jafar} shows that even global attention mechanisms learn to focus on nearby positions.
Building on this, \method{} employs a Cross-Scale Neighborhood Attention mechanism where each high-resolution query attends only to a compact neighborhood around its corresponding low-resolution location, aligning receptive fields across resolutions while keeping attention localized and efficient.

This design brings two key benefits.
First, by constraining attention to depend only on the input image $\mathbf{I}$, \method{} eliminates the need for semantic low-resolution features when computing attention keys. This decoupling enables direct transfer across different VFMs without retraining.
Second, restricting attention to local neighborhoods significantly reduces key–query interactions, achieving about 40\% fewer GFLOPs compared to JAFAR ~\cite{jafar} while maintaining high reconstruction quality and supporting larger upsampling ratios with a smaller memory footprint (\autoref{tab:caracteristics}).

\begin{table*}[t]
\centering
\resizebox{\textwidth}{!}{%
\begin{tabular}{@{} l @{\hspace{0.15cm}} c c c c c c | c || c c c c c | c @{}}
\toprule
& & \multicolumn{6}{c}{\textbf{Segmentation mIoU (Pascal VOC12) \citep{pascalvoc}}} 
& \multicolumn{6}{c}{\textbf{Depth $\delta_1 (\%)$ (NYUv2) \citep{silberman2012nyuv2}}} \\ 
\cmidrule(lr){3-8} \cmidrule(lr){9-14}
\textbf{Method} & \textbf{V.A} 
& \textbf{DINOv2-R} & \textbf{RADIOv2.5} & \textbf{Franca} & \textbf{DINOv3} & \textbf{$\Delta$ Mean} & \textbf{DINOv3-7B}
& \textbf{DINOv2-R} & \textbf{RADIOv2.5} & \textbf{Franca} & \textbf{DINOv3} & \textbf{$\Delta$ Mean} & \textbf{DINOv3-7B} \\ 
\midrule
Nearest & \cmark 
& 79.75 & 81.71 & 78.70 & 84.99 & --- & 85.37
& 79.47 & 82.49 & 79.89 & 85.25 & --- & 89.92 \\ 
\rowcolor{gray!25} FeatUp \citep{featup} & \xmark  
& 83.91 & 84.47 & 81.36 & 84.43 & \gain{+3.29} & \oomcell
& 82.34 & \sbest{85.31} & 80.67 & \best{87.11} & \gain{+2.09} & \oomcell \\ 
Bilinear & \cmark  
& 83.07 & 84.46 & 81.30 & 86.99 & \gain{+3.34} & \sbest{87.96}
& 81.58 & 83.90 & 81.20 & 86.10 & \gain{+1.78} & 90.69 \\ 
AnyUp \citep{wimmer2025anyup} & \cmark 
& 85.49 & 85.51 & 81.98 & 86.62 & \gain{+4.09} & 87.55
& \sbest{83.96} & 84.89 & \sbest{82.12} & 86.36 & \sbestgain{+2.52} & \sbest{91.24} \\ 
\rowcolor{gray!25} JAFAR \citep{jafar} & \xmark  
& \sbest{86.31} & \sbest{85.93} & \sbest{82.29} & \sbest{87.10} & \sbestgain{+5.12} & \oomcell
& 83.88 & 84.40 & 81.93 & 86.37 & \gain{+2.39} & \oomcell \\ 
\rowcolor{lightpastelblue} 
\method{} (ours) & \cmark 
& \best{86.46} & \best{86.60} & \best{83.07} & \best{87.85} & \bestgain{+5.58} & \best{88.84}
& \best{84.75} & \best{85.47} & \best{82.67} & \sbest{86.73} & \bestgain{+3.16} & \best{91.74} \\ 
\bottomrule
\end{tabular}%
}
\caption{\textbf{Semantic segmentation (mIoU $\uparrow$) and depth estimation ($\delta_1 \uparrow$)}: Results on Pascal VOC \citep{pascalvoc} and NYUv2 \citep{silberman2012nyuv2} using features from different VFMs: DINOv2-R \citep{darcet2023vitneedreg}, RADIOv2.5-B \citep{heinrich2025radiov25}, Franca-B \citep{venkataramanan2025franca}, DINOv3-B \citep{simeoni2025dinov3}. `\textbf{$\Delta$ Mean}' is computed against Nearest. We highlight \best{best} and \sbest{second best} scores, and \bestgain{best gain}. \textbf{V.A} indicates VFM-agnostic models. \oomcell{} indicates training `Out-of-Memory'.}
\label{tab:base_models}
\end{table*}

\paragraph{Analogy with Classical Filters}

Our architecture parallels classical joint filtering methods, such as Joint Bilateral Filtering and its upsampling variant (JBU) \citep{kopf2007jbu} formulated in \autoref{eq:filter}.
Indeed, the Cross-Scale Neighborhood Attention formulation can be directly interpreted as a two-component filtering process, with (1) \emph{Spatial Kernel}: product of RoPE encode relative positional information, serving as the localized spatial kernel, and (2) \emph{Content Kernel}: The dot-product attention weights between high-resolution queries and low-resolution keys define an adaptive content kernel, using similarity information derived solely from the input image.
Unlike approaches that use separate branches for query and key encoding \citep{jafar, wimmer2025anyup}, $\text{\method{}}$ derives keys directly from pooled, position-aware queries. This design guarantees that the guidance mechanism is independent of VFM features. In addition the local mechanism allows for each pixel to attend only to its nearby region, capturing spatial proximity and appearance similarity from the encoded guidance image.
Consequently, \method{} is fully VFM-agnostic during inference and remains substantially faster and more memory-efficient than full attention mechanisms.

\subsection{Training}
\label{sec:method:training}

\method{} is trained following a procedure similar to that of \citet{jafar}. 
Given a high-resolution input image $\mathbf{I}^{\mathrm{HR}}$, we generate a corresponding low-resolution image $\mathbf{I}^{\mathrm{LR}}$ by applying bilinear downsampling with a factor of~2.

Features extracted from $\mathbf{I}^{\mathrm{HR}}$ using a VFM serve as the ground-truth high-resolution features $\mathbf{F}^{\mathrm{HR}}$. 
Similarly, features extracted from $\mathbf{I}^{\mathrm{LR}}$ with the same VFM define the low-resolution inputs $\mathbf{F}^{\mathrm{LR}}$.
\method{} then upsamples $\mathbf{F}^{\mathrm{LR}}$ into $\widehat{\mathbf{F}}^{\mathrm{HR}} := \operatorname{NAF}(\mathbf{I^\mathrm{HR}, \mathbf{F}^\mathrm{LR}})$, using $\mathbf{I}^{\mathrm{HR}}$ as the guidance image.
For supervision, we employ a simple $\ell_2$ reconstruction loss between the predicted and ground-truth features:
$\mathcal{L}_{\text{train}} = \lVert\widehat{\mathbf{F}}^{\mathrm{HR}} - \mathbf{F}^{\mathrm{HR}}\rVert^2_2$.

Unlike previous works such as FeatUp~\citep{featup}, LoftUp~\citep{loftup}, or AnyUp~\citep{wimmer2025anyup}, we do not rely on additional regularization terms such as total variation, segmentation masks, or cropping consistency losses. 
This minimalist training setup highlights the robustness of \method{}, which achieves strong performance despite its simplicity. More details about the training are provided in \autoref{sec:supp_training_details}.

\section{Experiments}
\label{sec:experiments}

We evaluate the effectiveness of \method{} across multiple VFMs, tasks, and datasets.
First, in \autoref{sec:exp:linear_probe}, we assess its upsampling quality through linear probing on semantic segmentation and depth estimation.
Then, in \autoref{sec:exp:downstream}, we study the consistency and usefulness of the upsampled features for downstream applications such as open-vocabulary and video object segmentation. More details about the datasets and experiments are provided in \autoref{sec:supp_tasks_details}.

\subsection{Linear probing of upsampled features}
\label{sec:exp:linear_probe}

To assess the quality of our VFM-agnostic upsampler, we conduct linear probing experiments on semantic segmentation and depth estimation.
All VFMs take as input $448 \times 448$ images normalized according to their corresponding preprocessing. The extracted representations are upsampled to the original image resolution, corresponding to a $\times14$ as in \citep{venkataramanan2025franca,oquab2023dinov2} or $\times16$ as in \citep{heinrich2025radiov25,simeoni2025dinov3} spatial scaling depending on the VFM. A linear layer is then trained on top of the upsampled features to predict per-pixel semantic labels or depth values, depending on the task. For VFM-specific upsamplers \citep{featup,jafar}, models are trained on each corresponding VFM following official training codes before being frozen during probing. 

\subsubsection{Semantic segmentation.}

\paragraph{Across VFMs.}
We evaluate \method{} against both VFM-specific upsamplers \citep{featup,jafar} and VFM-agnostic approaches such as bilinear interpolation and AnyUp \citep{wimmer2025anyup}.  
To assess VFM generalization, we test all methods across a diverse set of strong vision foundation models: DINOv2-R-B \citep{darcet2023vitneedreg}, RADIOv2.5-B \citep{heinrich2025radiov25}, Franca-B \citep{venkataramanan2025franca}, DINOv3-B \citep{simeoni2025dinov3}, and the large-scale DINOv3-7B \citep{simeoni2025dinov3}.  
Experiments are conducted on Pascal VOC \citep{pascalvoc} and results are reported in \autoref{tab:base_models} (left). 

\method{} achieves the best results across all VFMs, with an average \bestgain{+5.58 mIoU} improvement over the `Nearest' baseline. The next-best upsampler, JAFAR \cite{jafar} reaches a {+5.12 mIoU} average gain, but it has to be retrained for each VFM. Notably, \method{} is the first VFM-agnostic approach to surpass VFM-specific models such as JAFAR, whereas AnyUp \citep{wimmer2025anyup} fails to do so.  
Moreover, VFM-dependent upsamplers cannot be trained on large models like DINOv3-7B due to memory constraints (even with batch size 1 on an A100 40GB-GPU), while \method{} remains applicable and continues to improve performance on linear probing (see. \autoref{tab:base_models}) and downstream tasks (see. \autoref{tab:downstream}).

\paragraph{Across datasets.}
We next evaluate \method{} on multiple semantic segmentation benchmarks: COCO \citep{lin2014coco}, Pascal VOC \citep{pascalvoc}, ADE20K \citep{fhou2017ade20k}, and Cityscapes \citep{cordts2016cityscapes}, while fixing the VFM to DINOv3-B \citep{simeoni2025dinov3}. Results are reported in \autoref{tab:seg_probing}.
In this setting, surprisingly, recent upsamplers such as JAFAR \citep{jafar}, FeatUp \citep{featup}, and AnyUp \citep{wimmer2025anyup} fail to outperform the bicubic baseline \citep{keys2003bicubic} while more classical methods such as Joint Bilateral Filtering or Joint Bilateral Upsampling lead to better scores. 

\method{} leads to the best results and consistently improves performance across all datasets, demonstrating strong robustness and generalization, with an average \bestgain{+4.23 mIoU} gain over the nearest-neighbor upsampling baseline, and a substantial improvement on Cityscapes for fine-grained segmentation. 

\begin{table}[t]
\centering
\resizebox{\columnwidth}{!}{%
\begin{tabular}{@{} l @{\hspace{0.15cm}} c c c c c c @{}}
\toprule
\textbf{Method} & \textbf{V.A} & \textbf{COCO} & \textbf{VOC} & \textbf{ADE20K} & \textbf{Cityscapes} & $\Delta$ \textbf{Mean} \\
\midrule
Nearest & \cmark & 63.78 & 84.99 & 44.23 & 58.73 & --- \\
\rowcolor{gray!20!} FeatUp \citep{featup} & \xmark & 64.24 & 87.07 & 46.10 & 62.58 & \gain{+1.57} \\
AnyUp \citep{wimmer2025anyup} & \cmark & 65.49 & 86.63 & 46.00 & 60.35 & \gain{+2.19} \\
Bilinear & \cmark & 65.65 & 86.99 & 46.37 & 63.08 & \gain{+2.59} \\
\rowcolor{gray!20!} JAFAR \citep{jafar}  & \xmark & 65.85 & 87.10 & \sbest{46.72} & 62.46 & \gain{+2.60} \\
Bicubic  & \cmark & 65.65 & 87.20 & 46.45 & 63.62 & \gain{+2.80} \\
JBF \citep{kopf2007jointbilateral} & \cmark & \sbest{65.91} & \sbest{87.24} & 46.68 & \sbest{63.85} & \sbestgain{+2.99} \\
JBU \cite{kopf2007jbu} & \cmark & 65.64 & 87.00 & 46.36 & 63.02  & \gain{+2.57} \\
\rowcolor{lightpastelblue}
\method{} (ours) & \cmark & \best{66.37} & \best{87.86} & \best{47.41} & \best{64.98} & \bestgain{+4.23} \\
\bottomrule
\end{tabular}%
}
\caption{\textbf{Semantic segmentation (mIoU $\uparrow$), using Dinov3-B \citep{simeoni2025dinov3} features, on various datasets}: COCO \cite{lin2014coco}, Pascal VOC \citep{pascalvoc}, ADE20K \citep{fhou2017ade20k}, and Cityscapes \citep{cordts2016cityscapes}. `\textbf{$\Delta$ Mean}' is computed against Nearest. We highlight \best{best} and \sbest{second best} scores, and \bestgain{best gain}. \textbf{V.A} indicates VFM-agnostic models.}
\label{tab:seg_probing}
\end{table}

\paragraph{Across model sizes.}
Finally, we study how upsampling methods scale with model size using the DINOv2-R \citep{darcet2023vitneedreg} family, including Small (S), Base (B), and Large (L) variants.  
Results are reported in \autoref{tab:seg_size_probing}.
\method{} again delivers consistent improvements across all sizes, achieving an average \bestgain{+5.27 mIoU} over the nearest baseline and setting new state-of-the-art scores at every size.  

To verify the generality of our findings, we further evaluate \method{} on a broader set of randomly selected VFM–dataset pairs (e.g., PE-Core \citep{bolya2025PerceptionEncoder}, CAPI \citep{darcet2025capi}, DINO \citep{caron2021dino}, PE-Spatial \citep{bolya2025PerceptionEncoder}, SigLIP2 \citep{tschannen2025siglip}, etc.) including model size from Tiny to Large. The detailed results are provided in \autoref{sec:supp_generalization}. The observed trends and performance gains remain consistent with those reported here, confirming the robustness of our conclusions. We also report scores for other and weaker VFM-specific upsamplers \citep{lift,featsharp}.

\subsubsection{Depth estimation}
We evaluate the quality of upsampled feature maps following the Probe3D \citep{banani2024probe3d} protocol, for depth estimation on the NYUv2 dataset \citep{silberman2012nyuv2}.  
Experiments are conducted across multiple base VFMs, and results are summarized in \autoref{tab:base_models} (right).  

{As opposed to semantic segmentation, depth estimation is a regression task which requires fine-grain prediction. Again,}
\method{} consistently achieves the best performance across all tested VFMs, with an average \bestgain{+3.16 $\delta_1$} improvement over the `Nearest' interpolation baseline and a +0.57 gain over AnyUp \citep{wimmer2025anyup}.  
Notably, \method{} also enhances the performance of the large-scale DINOv3-7B model \citep{simeoni2025dinov3}, yielding a substantial +12.69 $\delta_1$ increase compared to Nearest-neighbor interpolation.

\begin{table}[t]
\resizebox{\columnwidth}{!}{%
\begin{tabular}{@{} l @{\hspace{0.1cm}} c c c c c @{}}
    \toprule
    \textbf{Method} & \textbf{V.A} 
    & \textbf{DINOv2-R-S} 
    & \textbf{DINOv2-R-B} 
    & \textbf{DINOv2-R-L} 
    & \textbf{$\Delta$ Mean} \\
    \midrule
    Nearest & \cmark  & 38.61 & 40.73 & 41.00  & --- \\
    Bilinear & \cmark & 40.70 & 43.54 & 44.30  & \gain{+2.74} \\
    \rowcolor{gray!20!} FeatUp \citep{featup} & \xmark & 41.35 & 44.28 & 46.01 & \gain{+3.77} \\
    AnyUp \citep{wimmer2025anyup} & \cmark & 41.42 & 44.91 & 46.39 & \gain{+4.13} \\
    \rowcolor{gray!20!} JAFAR \citep{jafar} & \xmark & \sbest{42.13} & \sbest{45.64} & \sbest{47.04} & \sbestgain{+4.82}\\
    \rowcolor{lightpastelblue} 
    \method{} (ours) & \cmark &  \best{42.69} & \best{46.14} & \best{47.31} & \bestgain{+5.27} \\
\bottomrule
\end{tabular}%
}
\caption{\textbf{Semantic segmentation (mIoU $\uparrow$) on ADE20K \citep{fhou2017ade20k}, using features from DINOv2-R \citep{darcet2023vitneedreg} models of different sizes}: DINOv2-R-S, DINOv2-R-B, and DINOv2-R-L. `\textbf{$\Delta$ Mean}' is computed against Nearest. We highlight \best{best} and \sbest{second best} scores, and \bestgain{best gain}. \textbf{V.A} indicates VFM-agnostic models.}

\label{tab:seg_size_probing}
\end{table}

\subsection{Downstream transfer}
\label{sec:exp:downstream}

\begin{table*}[t]
\centering
\small
\resizebox{\textwidth}{!}{%
\begin{tabular}{l c ccccc | c || ccccc | c}
\toprule
& & \multicolumn{6}{c}{\textbf{Open Vocabulary Segmentation}} & \multicolumn{6}{c}{\textbf{Video Object Segmentation}} \\
& & \multicolumn{6}{c}{\small Pascal VOC \citep{pascalvoc}, mIoU (\%)} & \multicolumn{6}{c}{\small DAVIS \citep{ponttuset2017davis}, $\mathcal{J\&F}$ Mean} \\
\cmidrule(lr){3-8}\cmidrule(lr){9-14}
\textbf{Method} & \textbf{V.A} & \textbf{DINOv2-R} & \textbf{RADIO} & \textbf{Franca} & \textbf{DINOv3} & $\bm{\Delta}$ \textbf{Mean} & \textbf{DINOv3-7B} & \textbf{DINOv2-R} & \textbf{RADIO} & \textbf{Franca} & \textbf{DINOv3} & $\bm{\Delta}$ \textbf{Mean} & \textbf{DINOv3-7B} \\
\midrule
Bilinear (default) & \cmark & \sbest{62.00} & 47.07 & 60.95 & 62.21 & {0.00} & 61.88 & 64.36 & 63.75 & 70.52 & \sbest{70.00} & {0.00} & \sbest{69.50} \\
AnyUp~\cite{wimmer2025anyup} & \cmark & \sbest{62.00} & 46.49 & 62.72 & 63.41 & \gain{+0.60} & \sbest{62.88} & \sbest{69.56} & 68.16 & 68.90 & 65.90 & \gain{+0.97} & 69.03 \\
\rowcolor{gray!10} FeatUp~\cite{featup} & \xmark & 59.53 & \best{51.08} & 61.55 & 62.54 & \gain{+0.62} & \oomcell & 64.17 & 65.24 & \sbest{71.17} & 69.53 & \gain{+0.37} & \oomcell \\
\rowcolor{gray!10} JAFAR~\cite{jafar} & \xmark & \best{63.04} & 44.32 & \best{63.67} & \sbest{63.72} & \sbestgain{+0.63} &  \oomcell & \best{72.94} & \best{69.72} & 70.89 & 69.24 & \bestgain{+3.54} & \oomcell \\
\rowcolor{lightpastelblue} \textbf{\method{} (ours)} & \cmark & 60.69 & \sbest{48.96} & \sbest{62.88} & \best{63.86} & \bestgain{+1.04} & \best{63.06} & 69.49 & \sbest{69.25} & \best{72.82} & \best{70.55} & \sbestgain{+3.37} & \best{71.39} \\
\bottomrule
\end{tabular}
}
\caption{\textbf{Comparison of upsampling methods across downstream tasks.} using features from different VFMs: DINOv2-R \citep{darcet2023vitneedreg}, RADIOv2.5-B \citep{heinrich2025radiov25}, Franca-B \citep{venkataramanan2025franca}, DINOv3-B \citep{simeoni2025dinov3}. \textit{Left:} Open vocabulary segmentation using ProxyCLIP on Pascal VOC \citep{pascalvoc} (mIoU, \%). \textit{Right:} Video object segmentation propagation on DAVIS ($\mathcal{J\&F}$ Mean). All VFMs use Base (B) variants. `\textbf{$\Delta$ Mean}' is computed against Nearest. We highlight \best{best} and \sbest{second best} scores, and \bestgain{best gain}. \textbf{V.A} indicates VFM-agnostic models. \oomcell{} indicates training `Out-of-Memory'.}
\label{tab:downstream}
\end{table*}

We extend our evaluation by analyzing how upsampling affects the transferability and consistency of feature representations across different downstream tasks.  
Specifically, we investigate two complementary aspects: (i) the transferability of upsampled features to open-vocabulary segmentation, and (ii) their temporal consistency across video frames through video label propagation. %

\paragraph{Transfer to open-vocabulary segmentation.}
We first study whether spatial improvements from upsampling translate into better semantic transfer on open-vocabulary segmentation.  
We use ProxyCLIP \citep{lan2024proxyclip} to evaluate upsampled representations, replacing its default bilinear upsampling with different upsamplers including \method{} as a drop-in replacement without additional training. 

As reported in \autoref{tab:downstream} (left), \method{} achieves the highest average performance across evaluated VFMs, confirming its compatibility with many tasks. Moreover, while some upsamplers perform slightly better on specific VFM, \method{} consistently yields the best overall results, reaching a \bestgain{+1.04 mIoU} improvement over the baseline, compared to {+0.63 mIoU} for the second-best method, JAFAR \citep{jafar}.

\paragraph{Transfer to video segmentation.}
We next evaluate the temporal consistency of upsampled features through video segmentation propagation on the DAVIS dataset \citep{ponttuset2017davis}.  
Following the protocol of \citet{lift}, we extract dense features for each frame, upsample them, and propagate segmentation masks across frames using feature-space similarity matching. As in ProxyCLIP, we replace the default bilinear upsampling with different upsampler choices; our method can again be used as a drop-in replacement.

As reported in \autoref{tab:downstream} (right), \method{} yields the best overall performance, achieving an average \sbestgain{+3.37 mIoU} improvement over the baseline, highlighting the effectiveness of our approach to keep feature consistency through frames.  

\section{Ablations}
\label{sec:ablations}

\begin{table}[b]
\centering
\resizebox{\columnwidth}{!}{%
\begin{tabular}{@{}l @{\hspace{0.2cm}}c@{\hspace{0.25cm}}c@{\hspace{0.25cm}}c@{\hspace{0.25cm}}c @{\hspace{0.25cm}}c | @{\hspace{0.15cm}}c@{\hspace{0.15cm}}c@{}}
\toprule
 & \hspace{-0.2cm}\textbf{DINOv2-R} & \textbf{RADIOv2.5} & \textbf{Franca} & \textbf{DINOv3} & \textbf{Mean} & \textbf{Params (M)} & \textbf{FPS} \\ 
\midrule
\textit{\textbf{Encoders}} & & & & & & & \\
\quad Pixel Enc. & 62.87& 58.13 & 57.46 &60.96 & 59.86 & 0.26 & 19 \\
\rowcolor{lightpastelblue} \quad + Context Enc. & 63.82 & 58.76& 58.03&61.01 & \textbf{60.41} & 0.66 & 18 \\
\midrule
\textit{\textbf{Block Type}} & & & & & & & \\
\quad Inception~\cite{szegedy2015inception} & 62.73 &57.97 &57.17 &60.37 & 59.56 & 0.15 & 15 \\
\quad ResNet & 62.99 &58.15 &57.52 &60.49 & 59.79 & 0.66 & 18 \\
\rowcolor{lightpastelblue} \quad Dual-Branch & 63.82 & 58.76 & 58.03 & 61.01 & \bf 60.41 & 0.66 & 18 \\
\midrule
\textit{\textbf{Guidance dim.}} & & & & & & & \\
\quad $C = 64$ & 62.90&57.89 &57.04 &60.14 & 59.49 & 0.04 & 40 \\
\quad $C = 128$ & 63.49&58.48 &57.46 &60.59 & 60.01 & 0.16 & 30 \\
\rowcolor{lightpastelblue} \quad $C = 256$ & 63.82 & 58.76& 58.03&61.01 & 60.41 & 0.66 & 18 \\
\quad $C = 512$ & 63.98& 58.86&58.33 &61.10 & 60.57 & 2.63 & 9 \\
\quad $C = 768$ & 64.23 & 59.11 & 58.49 & 61.47 & 60.82 & 5.92 & 6  \\
\quad $C = 1024$ & 64.45 & 59.41 & 58.63 & 61.73 & \textbf{61.05} & 10.5 & 4 \\
\midrule
\textit{\textbf{\# conv.\ blocks}} & & & & & & & \\
\quad $L = 1$ & 62.97&58.14 &57.55 & 60.41 & 59.77 & 0.33 & 27 \\
\rowcolor{lightpastelblue} \quad $L = 2$ & 63.82 & 58.76& 58.03&61.01 & 60.41& 0.66 & 18 \\
\quad $L = 3$ & 64.04& 58.92&58.24 &61.15 & 60.59 & 0.99 & 14 \\
\quad $L = 4$ & 64.54 & 59.27 & 58.61 & 61.88 & 61.08 & 1.32 & 11 \\
\quad $L = 5$ & 64.55 & 59.34 & 58.81 & 61.82 & \textbf{61.13} & 1.65 & 9\\
\midrule
\rowcolor{Apricot!20!} \method{}++ & 69.94 & 61.51 & 61.10 & 65.69 & \textbf{64.56} & 14.8 & 3 \\

\bottomrule
\end{tabular}%
}
\caption{\textbf{Ablation of the dual-branch image encoder: mIoU (\% $\uparrow$)} on Cityscapes. \textcolor{lightpastelblue}{Blue} and \textcolor{Apricot!}{orange} rows highlight the base configuration and the best one. VFMs are of the Base (B) variant.}
\label{tab:ablation_block_design}
\end{table}

\paragraph{Design of the Dual-Branch Encoder.}
We first analyze the design of the dual-branch guidance encoder $\operatorname{Enc}_\theta$. The ablation studies isolate the impact of four factors: (1) the presence of the context branch (3$\times$3 conv) in addition to the pixel branch (1$\times$1 conv), (2) the block design (our separate dual-branch blocks vs. ResNet~\cite{he2016deep} and Inception-style blocks~\citep{szegedy2015inception}), (3) the output guidance dimension $C$, and (4) the number of stacked blocks $L$.
Results are reported in \autoref{tab:ablation_block_design}, where the configuration used in all other experiments is highlighted in light blue.
Key observations are as follows:
\begin{itemize}
    \item Adding the context branch is crucial for capturing local structure beyond per-pixel information.
    \item Our dual-branch encoder outperforms Inception-style blocks while remaining simpler and more efficient.
    \item Increasing $C$ and $L$ improves accuracy but with diminishing returns and higher computational cost.
\end{itemize}

\begin{table*}[ht]
\centering
\resizebox{\textwidth}{!}{%
\begin{tabular}{@{} l cccc c | l cccc c @{}}
\toprule
 & \textbf{DINOv2-R} & \textbf{RADIOv2.5} & \textbf{Franca} & \textbf{DINOv3} & \textbf{Mean} 
&  & \textbf{DINOv2-R} & \textbf{RADIOv2.5} & \textbf{Franca} & \textbf{DINOv3} & \textbf{Mean} \\
\midrule
\textit{\textbf{Pooling design for attention keys $K$}} \hspace{-10cm} & & & & & & 
\textit{\textbf{Spatial encoding design}} \hspace{-5cm} \\ %
AvgPool + Conv. & 61.08 & 56.63 & 56.08 & 59.71 & 58.38 & $\emptyset$ & 60.32 & 54.91 & 55.52 & 57.52 & 57.07 \\
Bilinear & 62.79 & 57.76 & 57.27 & 60.80 & 59.66 
& Manhattan & 61.47 & 55.68 & 56.48 & 58.47 & 58.03 \\
MaxPool & 63.35 & 58.45 & 57.68 & 60.25 & 59.93 
& Gaussian & 61.65 & 55.87 & 56.52 & 58.71 & 58.19 \\
\rowcolor{lightpastelblue} AvgPool & \textbf{63.82} & \textbf{58.76} & \textbf{58.03} & \textbf{61.01} & \textbf{60.41} &
RoPE & \textbf{63.82} & \textbf{58.76} & \textbf{58.03} & \textbf{61.01} & \textbf{60.41} \\
\bottomrule
\end{tabular}%
}
\caption{\textbf{Ablations on the design of the attention keys $K$ and spatial encoding.} We report linear probing semantic segmentation results on Cityscapes \citep{cordts2016cityscapes} (mIoU, \%, $\uparrow$).
The setting used by NAF is in \textcolor{lightpastelblue!}{blue}. All VFMs are of the Base (B) variant.}
\label{tab:ablation_spatial_range}
\end{table*}

For all main experiments, we adopt a lightweight configuration with $C=256$ and $L=2$, that has roughly the same number of parameters of previous state-of-the art VFM-specific upsampler \citep{jafar}. It offers a good trade-off between performance and efficiency.
A larger variant, NAF++ shown in \autoref{tab:ablation_block_design}, uses higher $C=768$ and $L=5$ for improved accuracy at the cost of additional parameters and lower FPS. %

\paragraph{Design of the attention keys $K$.}
We evaluate alternative designs for the attention keys $K$ used in the filtering process.
Our default choice, `AvgPool' defined in \autoref{eq:keys}, averages the guidance features within each low-resolution region. We compare this to three variants: (1) `MaxPool': replaces average pooling with a max operation, (2) `AvgPool + Conv.': applies a $1\times1$ convolution after average pooling to mix channel information, similar to the design used in JAFAR \citep{jafar} and AnyUp \citep{wimmer2025anyup}, (3) {`Bilinear': removes pooling entirely and computes $K_q$ by bilinear interpolation.}

Results in \autoref{tab:ablation_spatial_range} (left) show that pooling is essential: both `AvgPool' and `MaxPool' outperform the Bilinear variant, confirming the need for local aggregation.
However, adding a convolution on top of pooled features, mixing independently channels from queries and keys as done in previous works \citep{jafar,wimmer2025anyup}, significantly degrades performance, likely by breaking the channel alignment between queries and keys. We therefore adopt the simple `AvgPool' design in all experiments.

\begin{table}[b]
\centering
\resizebox{\columnwidth}{!}{%
\begin{tabular}{@{} l r cccccc @{}}
\toprule
& & \multicolumn{6}{c}{\textbf{Gaussian Denoising} ({PSNR} / {SSIM} (\%))} \\
\textbf{Method} & \hspace{-3cm} \# Params (M) & \multicolumn{2}{c}{\textbf{$\sigma=0.1$}} & \multicolumn{2}{c}{\textbf{$\sigma=0.5$}} & \multicolumn{2}{c}{\textbf{$\sigma \in [0.1, 0.5]$}} \\
\cmidrule(lr){1-2}\cmidrule(rl){3-4} \cmidrule(rl){5-6} \cmidrule(rl){7-8}
DnCNN \citep{zhang2017dncnn} & 0.56 & 30.86 & 87.1  & 22.82 & 57.3 & 20.23 & 40.5 \\
IRCNN \citep{zhang2017ircnn} & 0.19 & 31.39 & 88.9 & 23.87 & 65.0 & 25.86 & 73.6 \\
REDNet \citep{jiang2018rednet} & 1.04 & 31.99 & 89.8 & 24.46 & 67.4 & 26.52 & 74.6 \\
Restormer \citep{Zamir2021Restormer} & 26.13 & \best{32.51} & \sbest{90.5} & \best{25.51} & \best{73.2} & \best{27.77} & \best{79.9} \\
\rowcolor{lightpastelblue}\method{} (ours) & 0.66 & \sbest{32.12} & \best{90.9} & \sbest{24.52} & \sbest{68.8} & \sbest{27.11} & \sbest{77.2} \\
\midrule
& & \multicolumn{6}{c}{\textbf{Channel-Wise Salt \& Pepper Denoising} ({PSNR} / {SSIM} ($\%$))}  \\
\textbf{Method} & \hspace{-3cm} \# Params (M) & \multicolumn{2}{c}{\textbf{$p=0.1$}} & \multicolumn{2}{c}{\textbf{$p=0.5$}} & \multicolumn{2}{c}{\textbf{$p \in [0.1, 0.5]$}} \\
\cmidrule(lr){1-2}\cmidrule(rl){3-4} \cmidrule(rl){5-6} \cmidrule(rl){7-8}
DnCNN \citep{zhang2017dncnn} & 0.56 & 35.62 & 98.2 & 28.44 & 81.0 & 27.72 & 87.3 \\
IRCNN \citep{zhang2017ircnn} & 0.19 & 35.09 & 97.1 & 22.75 & 62.3 & 25.07 & 71.7 \\
REDNet \citep{jiang2018rednet} & 1.04 & 35.14 & 95.7 & 25.11 & 73.9 & 26.46 & 75.3 \\
Restormer \citep{Zamir2021Restormer} & 26.13 & \best{47.76} & \sbest{99.6} & \best{36.90} & \best{97.0} & \best{40.41} & \sbest{98.2} \\
\rowcolor{lightpastelblue}\method{} (ours) & 0.66 & \sbest{47.47} & \best{99.7} & \sbest{32.91} & \sbest{94.3} & \sbest{39.62} & \best{98.9} \\
\bottomrule
\end{tabular}%
}
\caption{\textbf{Gaussian and Channel-Wise Salt \& Pepper Denoising (PSNR $\uparrow$ / SSIM $\uparrow$)} on ImageNet.  
The standard deviation for gaussian noise $\sigma$, and the channel-wise salt-and-pepper noise corruption probability $p$ are either fixed to $0.1$ or $0.5$, or randomly sampled from $[0.1, 0.5]$.}
\label{tab:denoising}
\end{table}

\paragraph{Spatial positional encoding.}
To enable spatial relationship reasoning in the attention formulation, $\text{\method{}}$ uses positional embeddings on its guidance features. Our default strategy is RoPE, applied to both queries and keys (\autoref{eq:queries}–\autoref{eq:keys}), which encodes relative spatial offsets directly into the attention computation. Beyond the following ablation, a more in depth mathematical analysis motivating the choice of RoPE can be found in \autoref{sec:supp_maths}.
Yet we compare this to explicit multiplicative spatial kernels:
\begin{equation}
\langle Q_p, K_q \rangle =
\exp \Big(-\tfrac{|p - q|\ast}{2\sigma^2}\Big)
\big\langle
\operatorname{Enc}_\theta(\mathbf{I})_p,
\operatorname{Enc}_\theta(\mathbf{I})_q
\big\rangle,
\end{equation}
where $|\cdot|\ast$ is $|\cdot|_2^2$ (`Gaussian') or $|\cdot|_1$ (`Manhattan') and $\sigma$ is learnable.
We also test a variant without positional encoding (`$\emptyset$').
As shown in \autoref{tab:ablation_spatial_range} (right), positional encoding is crucial for spatial awareness.
RoPE achieves the best performance, efficiently capturing relative geometry without additional parameters.

\section{Extension: Image Restoration}

We extend \method{} beyond feature upsampling to demonstrate its broader applicability.  
This generalization follows naturally from its filter-based formulation: by design, \method{} imposes no constraints on the dimensionality or structure of its input and guidance signals.  
As a result, the same architecture can process either low-resolution feature maps (for upsampling) or standard RGB images (for restoration).

To illustrate this flexibility, we evaluate \method{} on image restoration tasks, where both the guidance and input correspond to the same image.  
Since there is no downsampling in this setting, the average pooling operation becomes the identity, and the query and key representations simplify to:
\begin{equation}
K = Q = \operatorname{RoPE}(\text{Enc}_\theta(\mathbf{I})).
\end{equation}

\smallskip\noindent\textbf{Task and setup.}
We focus on image denoising, where the goal is to recover a clean RGB image of size $448 \times 448$ from its corrupted version.  
Input images are generated by adding artificial noise to ground-truth samples.  
We consider two types of corruption:  
(1) additive Gaussian noise with standard deviation $\sigma$, and  
(2) channel-wise salt-and-pepper noise, meaning we randomly activate or desactivate some channels with corruption probability $p$.  
And we consider both fixed and dynamic noises.

We use the same architecture as in the feature upsampling experiments, except that we enlarge the neighborhood attention kernel to better capture spatial dependencies (from 9 to 15).  
The network is trained end-to-end with a combination of $\mathcal{L}_1$, $\mathcal{L}_2$, and SSIM losses as done in classical training pipelines. More details can be found in \autoref{sec:supp_restoration}.

\smallskip\noindent\textbf{Baselines.}
For a fair comparison, we retrain classical denoising networks~\citep{jiang2018rednet,zhang2017dncnn,zhang2017ircnn} and a state-of-the-art transformer-based model, Restormer~\citep{Zamir2021Restormer}, using the same data, losses, and training schedule.

\smallskip\noindent\textbf{Results.}
As shown in \autoref{tab:denoising}, \method{} achieves strong results across all noise levels and types, despite not being specifically designed for denoising.  
While Restormer remains the top performer, it relies on roughly $40\times$ more parameters and an architecture specifically designed for image restoration.  
Notably, conventional models perform well on Gaussian noise but degrade significantly under salt-and-pepper corruption, whereas \method{} effectively recovers structural details in both cases.  
These results confirm that the neighborhood attention mechanism, guided only by the image itself, generalizes well beyond feature upsampling to broader image restoration tasks.

\section*{Conclusion}

We introduced \method{}, a VFM-agnostic upsampling module capable of scaling any feature to any resolution, including very large 7B-VFMs, achieving state-of-the-art results on many downstream tasks. 
At its core lies a Cross-Scale Neighborhood Attention mechanism that eliminates dependency on the target feature distribution, drawing strong analogies to classical Joint Bilateral Filtering. Its local interpolation design allows it to be fast and lightweight compared to previous VFM upsamplers, enabling previously unseen scaling ratios. 
Finally, its simple yet powerful approach serves as a versatile module which can be used in many different tasks paving the way to broader applications.

\vspace{0.2em}
\section*{Acknowledgments}

We thank Amaia and Nicolas for proofreading the paper, and Yihong for constant support throughout the project.

{
    \small
    \bibliographystyle{ieeenat_fullname}
    \bibliography{biblio}
}

\clearpage
\setcounter{page}{1}
\maketitlesupplementary

\appendix

\section*{Table of contents}
\startcontents
\printcontents{ }{1}{}

\section{Mathematical Discussions}
\label{sec:supp_maths}

We analyze the mathematical foundations of \method{}’s upscaling mechanism, focusing on the interaction between RoPE \citep{su2024roformer} and neighborhood attention. Our key finding is that \method{} does not merely reweight the inputs using a distance over the image encoder; instead, it learns the Inverse Discrete Fourier Transform (IDFT) of the upsampling aggregation kernel. In other words, \method{} dynamically constructs an optimal upsampling filter by predicting spectral coefficients of the learned image encoder.

\paragraph{Preliminaries}
To recall, \method{} shows analogies with Joint Bilateral Filtering due to the spatial-and-content aware kernel. It allows to obtain high-resolution features via the following attention-weighted interpolation:
\begin{equation}
\mathbf{F}^{\mathrm{HR}}_p
= \frac{1}{Z(p)} \sum_{q \in \mathcal{N}(p)}
\exp\Big(\frac{1}{\sqrt{d}} S(p,q)\Big) \mathbf{F}^{\mathrm{LR}}_q,
\end{equation}
where $Z(p) = \sum_{q \in \mathcal{N}(p)} \exp\big(\frac{1}{\sqrt{d}} S(p,q)\big)$ is the normalization constant and $S(p,q) := \langle Q_p , K_q \rangle$ is an attention-score for a pair of points $(p,q)$ with queries and keys defined as
\begin{equation}
Q_{p} := \operatorname{RoPE}(\mathbf{G})_p,
\quad
K_{q} := \operatorname*{AvgPool}_{q' \in q}
\big[ \operatorname{RoPE}(\mathbf{G})_{q'} \big],
\label{eq:queries_keys}
\end{equation}
and $\mathbf{G} = \text{Enc}_\theta(\mathbf{I}) \in \mathbb{R}^d$ denotes the image encoder output having $d$ channels.

Substituting the pooled key yields the following attention-score:
\begin{equation}
\label{eq:score_pool_expansion}
S(p,q) =
\frac{1}{|\{q' \in q\}|} \sum_{q' \in q }
\langle \operatorname{RoPE}(\mathbf{G})_p , \operatorname{RoPE}(\mathbf{G})_{q'} \rangle.
\end{equation}

\subsection{RoPE Expansion}

\paragraph{RoPE introduction.}
To develop the last equation we discuss about 2D-RoPE \citep{su2024roformer}. To do so, we consider channel pairs $(2c, 2c+1)$ where $c \in \{0,...,d/2\}$ and define the 2D feature vector for the $c$-th pair as:
\begin{equation}
\vec{G}_c(p) :=
\begin{pmatrix} \mathbf{G}_p^{2c}\\[2pt] \mathbf{G}^{2c+1}_p \end{pmatrix},
\end{equation}
where $\mathbf{G}^{2c}_p$ is the value of the encoded image $\mathbf{G}$, at position $p$ and in channel $2c$.

By definition of RoPE we have for each $c$:
\begin{equation}
\operatorname{RoPE}(\vec{G}_c(p)) = R_c(p) \, \vec{G}_c(p),
\end{equation}
where the rotation matrix is
\begin{equation}
R_c(p) :=
\begin{pmatrix}
\cos \Phi_c(p) & -\sin \Phi_c(p) \\
\sin \Phi_c(p) & \cos \Phi_c(p)
\end{pmatrix}.
\end{equation}

with the rotation angle $\Phi_c(p)$ encoding the axial positional information for channel pair $c$. It is defined by:
\begin{equation}
\Phi_c(p) =
\begin{cases}
2\pi \, p_y / \lambda_c & \text{if } 0 \le c < d/4 \quad (\text{Height}) \\[2pt]
2\pi \, p_x / \lambda_c & \text{if } d/4 \le c < d/2 \quad (\text{Width})
\end{cases},
\end{equation}
with $\lambda_c$ the wavelength of the $c$-th frequency band, and $p_x, p_y \in [-1,1]$ are normalized coordinates along each axis.

\paragraph{Inner product after rotation.}  
We can reinject the definition of RoPE \citep{su2024roformer} in the attention-score between two positions $p$ and $q'$. It becomes:
\begin{equation}
\resizebox{\columnwidth}{!}{$
\langle \operatorname{RoPE}(\vec{G}_c(p)), \operatorname{RoPE}(\vec{G}_c(q')) \rangle
= \vec{G}_c(p)^\top R_c(p)^\top R_c(q') \, \vec{G}_c(q').
$}
\end{equation}

Using properties of 2D rotation matrices, the product $R_c(p)^\top R_c(q')$ is itself a rotation by the relative angle
\begin{equation}
\Delta \Phi_c(p,q') := \Phi_c(q') - \Phi_c(p).
\end{equation}
Hence, we can write
\begin{equation}
R_c(p)^\top R_c(q') =
\begin{pmatrix}
\cos \Delta \Phi_c(p,q') & -\sin \Delta \Phi_c(p,q') \\
\sin \Delta \Phi_c(p,q') & \cos \Delta \Phi_c(p,q')
\end{pmatrix}.
\end{equation}

\begin{figure}[t]
    \centering
    \includegraphics[width=\linewidth]{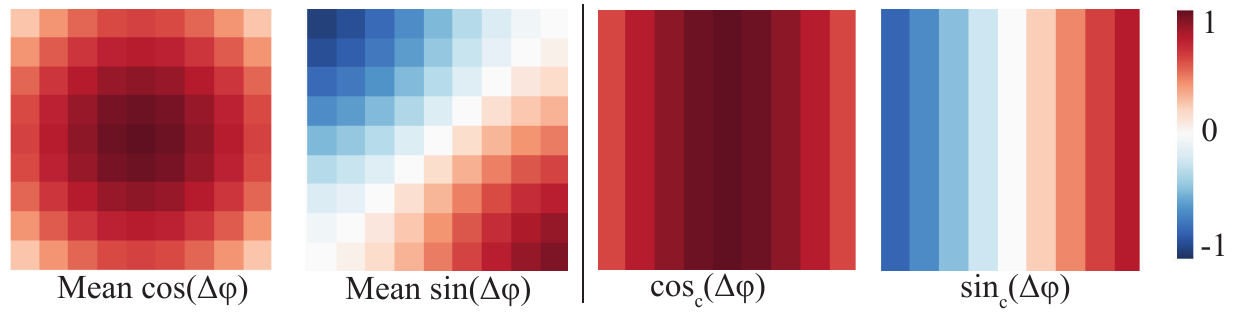}
    \caption{\textbf{Illustration of the mean and channel-specific cosine and sine of $\Delta \Phi_c$.}
We compute the mean across all channels and select a single random channel to illustrate its individual behavior. For the cosine, we observe an overall decreasing pattern as the distance from the center increases.}
    \label{fig:angles}
\end{figure}

To better visualize $\Delta \Phi_c$, we plot in \autoref{fig:angles} the mean of cosine and sine over all channels and their values at a specific channel given a neighborhood window of size $9 \times 9$. We see that in average the cosine decreases when the points become far, while the sine has a diagonal shape due to the axial-nature of $\Phi_c(p)$.

\paragraph{Dot and cross product decomposition.}
Expanding the inner product of 2D vectors under a rotation gives the per-channel contribution:
\begin{equation}
\label{eq:pairwise_Ac}
\begin{split}
A_c(p,q') &= \vec{G}_c(p)^\top R_c(p)^\top R_c(q') \, \vec{G}_c(q') \\
&= (\vec{G}_c(p) \cdot \vec{G}_c(q')) \cos \Delta \Phi_c(p,q') \\
&\quad - (\vec{G}_c(p) \times \vec{G}_c(q')) \sin \Delta \Phi_c(p,q'),
\end{split}
\end{equation}
with the standard dot and cross products.

\begin{figure}[ht!]
    \centering
    \includegraphics[width=\linewidth]{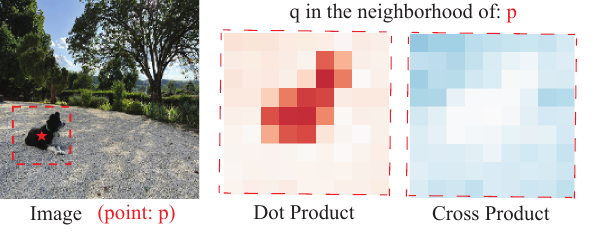}
    \caption{\textbf{Dot and cross products for a specific channel} given a query point p on an image. We highlight the neighborhood around p using a dashed red square. On the feature side, after VFM-downsampling, we observe that implicitly  \method{} discriminates boundaries.}
    \label{fig:cross_dot_weights}
\end{figure}

\paragraph{Interpretation.}  
The dot product $\vec{G}_c(p) \cdot \vec{G}_c(q')$ measures \textit{feature coherence} (alignment), while the cross product $\vec{G}_c(p) \times \vec{G}_c(q')$ captures \textit{content orthogonality} (perpendicularity). RoPE modulates these content interactions based strictly on the relative axial distance: vertical distance for $d/2$ channels and horizontal distance for the remaining $d/2$ channels. As we can see in \autoref{fig:cross_dot_weights}, the model learns to discriminate regions based on their encoding. While querying the dog, we recognize its shape and the model learns to aggregate inside values while discriminating outside ones.

\subsection{Representation with magnitudes and angles.}  
Let $\Psi_c(p,q')$ be the angle from $\vec{G}_c(p)$ to $\vec{G}_c(q')$, and let
\begin{equation}
r_p^{(c)} := \|\vec{G}_c(p)\|, \qquad r_{q'}^{(c)} := \|\vec{G}_c(q')\|
\end{equation}
denote the magnitudes of the feature vectors for channel pair $c$.

Then the dot and cross products can be expressed as:
\begin{equation}
\begin{split}
\vec{G}_c(p) \cdot \vec{G}_c(q') &= r_p^{(c)} r_{q'}^{(c)} \cos \Psi_c(p,q'), \\
\vec{G}_c(p) \times \vec{G}_c(q') &= r_p^{(c)} r_{q'}^{(c)} \sin \Psi_c(p,q'),
\end{split}
\end{equation}
so that the per-channel contribution becomes
\begin{equation}
\begin{split}
A_c(p,q') &= (\vec{G}_c(p) \cdot \vec{G}_c(q')) \cos \Delta \Phi_c(p,q') \\
& \qquad \qquad \quad - (\vec{G}_c(p) \times \vec{G}_c(q')) \sin \Delta \Phi_c(p,q') \\
&= r_p^{(c)} r_{q'}^{(c)} \big[ \cos \Psi_c(p,q') \cos \Delta \Phi_c(p,q') \\
&\qquad\qquad\quad - \sin \Psi_c(p,q') \sin \Delta \Phi_c(p,q') \big] \\
&= r_p^{(c)} r_{q'}^{(c)} \cos \big(\Psi_c(p,q') + \Delta \Phi_c(p,q')\big),
\end{split}
\end{equation}
where the last equality follows from the cosine angle addition formula. Finally, the pooled attention-score aggregates pairwise interactions over the pooling neighborhood as:
\begin{equation}
\resizebox{\columnwidth}{!}{$
\begin{split}
S(p,q)
&= \frac{1}{|\{q' \in q\}|}
\sum_{q' \in q}
\langle \operatorname{RoPE}(\mathbf{G})_p, \operatorname{RoPE}(\mathbf{G})_{q'} \rangle \\
&= \frac{1}{|\{q' \in q \}}
\sum_{q' \in q}
\sum_{c=0}^{d/2-1}
r_p^{(c)} r_{q'}^{(c)} \cos\!\big(\Psi_c(p,q')+\Delta\Phi_c(p,q')\big).
\end{split}
$}
\end{equation}

This decomposition clarifies the geometric mechanism of RoPE \citep{su2024roformer}. Rather than linearly adding a positional bias to a content score, \autoref{eq:score_pool_expansion} shows that position and content are \textbf{coupled} via phase addition.
The magnitude term $r_p^{(c)} r_q^{(c)}$ represents the raw signal strength (feature confidence). The cosine term indicates that the spatial phase difference $\Delta\Phi_c$ acts as a rotation applied to the semantic phase alignment $\Psi_c$. Constructive interference (a high score) occurs only when the semantic relationship compensates for the spatial offset, effectively implementing a spatially-varying matched filter.

\paragraph{Fourier-inspired Interpretation}

The derivation of the pairwise attention-score in Eq.~\eqref{eq:score_pool_expansion} reveals a structural equivalence to the Inverse Discrete Fourier Transform (IDFT). To see this, consider the standard reconstruction of a 1D spatial signal $f(x)$ from its frequency components:
\begin{equation}
    f(x) = \sum_{\omega} \underbrace{|F(\omega)|}_{\text{Amplitude}} \cdot \cos( \underbrace{\psi}_{\text{Content Phase}} + \underbrace{\omega \Delta x}_{\text{Spatial phase}}).
\end{equation}

\begin{figure}[t]
    \centering
    \includegraphics[width=\linewidth]{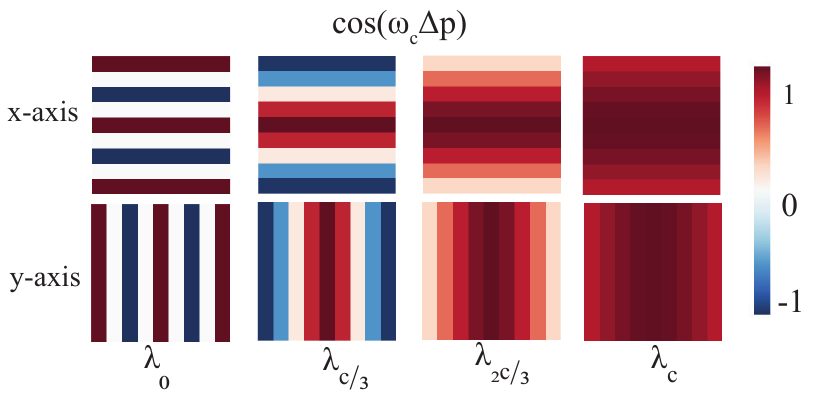}
    \caption{\textbf{Illustration of radial wavelets induced by \method{}} for a $9\times9$ neighborhood. We plot $\cos( \omega_c \cdot \Delta x)$ for different $\lambda$ where $\Delta x$ is defined as the $\ell_1$ distance over $x$-axis or $y$-axis between two coordinates over a grid map. In this plot we set the periods as: $\lambda_{i} = 100^{i / c}$.}
    \label{fig:wavelet}
\end{figure}

By viewing the channel dimension $c$ through the lens of Rotary Embeddings, where each channel corresponds to a specific wavelength $\lambda_c$, we can identify $c$ as a spectral frequency index $\omega_c \propto 1/\lambda_c$. We illustrate the resulting cosine and sine in \autoref{fig:wavelet}. Consequently, our derived attention-score $S(p, q')$ represented as a 1D score: $S(x)$ acts for each axis as a kernel synthesized via IDFT:
\begin{equation}
    S(x) \propto 
    \sum_{c} 
    \underbrace{r_p^{(c)} r_{q'}^{(c)}}_{\text{Amplitude}}
    \cos\!\big(
        \underbrace{\Psi_c}_{\text{Content Phase}}
        +
        \underbrace{\omega_c \Delta x}_{\text{Spatial Phase}}
    \big).
\end{equation}

This mapping offers three fundamental insights into the mechanism of \method{}:

\paragraph{1. Learning Fourier Coefficients.} 
The network does not directly predict spatial filter weights. Instead, it predicts the \textbf{Fourier series coefficients} of the optimal upsampling kernel. The product of feature magnitudes $ r_p^{(c)} r_{q'}^{(c)}$ acts as the \textit{spectral power} for frequency band $c$. By modulating these magnitudes, the encoder determines how much contribution each frequency—low (global structure) or high (fine detail)—makes to the final interpolation kernel.

\begin{figure}[ht!]
    \centering
    \includegraphics[width=\linewidth]{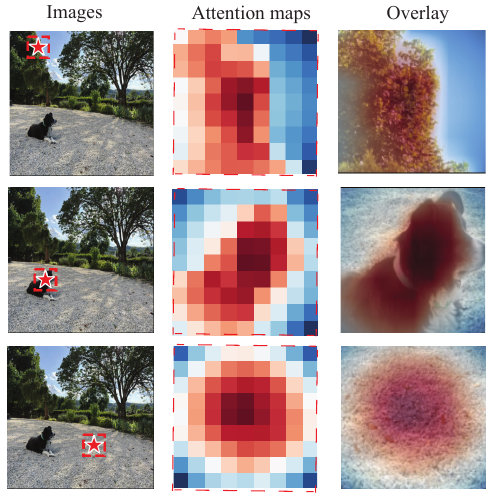}
    \caption{\textbf{Attention maps: $\langle Q_p , K_q \rangle$ between a query point $p$ and its $9 \times 9$ patch neighborhood ($q$)}. We take $896\times896$ input images to visualize finer details. In the first row, we see that \method{} learns to discriminate between the sky and the tree, i.e., borders. On the second row, it learns to discriminate more complex shapes (dog). On the third row, in uniform region, it shows decreasing attention pattern, akin to the gaussian filter used in classical JBF.}
    \label{fig:attention_maps}
\end{figure}

\paragraph{2. Spatially-Varying Filter Synthesis.} 
This formulation allows \method{} to function as a spatially-varying band-pass filter. In smooth image regions, the encoder can suppress high-frequency channels (reducing $\mathcal{A}_c$ for large $c$), effectively synthesizing a broad, low-pass smoothing kernel. Conversely, at sharp boundaries, the encoder can boost high-frequency amplitudes to synthesize a peaked, detail-preserving kernel (see \autoref{fig:attention_maps}).

\paragraph{3. Shift Demodulation.} 
The RoPE term $\omega_c \Delta x$ explicitly encodes the spatial shift theorem. By decomposing the interaction into spectral bands, the attention mechanism can align features that are semantically coherent but spatially phase-shifted. The summation over channels then reconstructs the spatial impulse response required to interpolate the feature at the exact sub-pixel position required by the target resolution.

\section{Feature Upsampling Experiments}

\subsection{Training Details}
\label{sec:supp_training_details}

\method{} is initially trained for 25k iterations with a batch size of 2, using input and target features extracted from $256 \times 256$ and $512 \times 512$ images, respectively, corresponding to a $\times 2$ upsampling. Subsequently, the module undergoes an additional 2.5k iterations (10\% of the initial training) using $1024 \times 1024$ images for the target features, while input features are drawn from images of varying resolutions between $256 \times 256$ and $896 \times 896$. Quantitatively, we observe an average of +0.4 mIoU gains on linear probing semantic segmentation evaluated on VOC \citep{pascalvoc}. Ablation studies in \autoref{sec:ablations} are performed without this second training stage.

The full training procedure requires approximately 1 hour and 9 GB of memory on an A100 GPU, resulting in a $\times 5$ speedup compared to the concurrent AnyUp method \citep{wimmer2025anyup}. For the final model, the dual-branch encoder employs $L=2$ blocks with $C=256$ channels and a neighborhood kernel size of 9. This configuration ensures a fair comparison with other upsamplers, maintaining a parameter count similar to JAFAR \citep{jafar} and an equivalent number of encoder blocks. For the neighborhood attention module, we adopt an efficient implementation \citep{hassani2022dilated,hassani2023neighborhood,hassani2024faster,hassani2025generalized}.

\subsection{Task setups}
\label{sec:supp_tasks_details}
\paragraph{Semantic Segmentation.}
To evaluate the upsamplers, we freeze their parameters and train linear classifiers on the extracted features following \citet{jafar}.
We train for 20 epochs on Cityscapes \citep{cordts2016cityscapes}, Pascal VOC \citep{pascalvoc}, and ADE20K \citep{fhou2017ade20k}, and for 5 epochs on COCO \citep{lin2014coco}.
We employ the AdamW optimizer with a learning rate of $5\times10^{-4}$ for most datasets; however unlike \citet{jafar}, for Cityscapes, we reduce the learning rate to $1\times10^{-4}$ to ensure stability.
A one-cycle cosine annealing scheduler is applied, and all input and target images are resized to $448\times448$.
The classifiers are optimized using the cross-entropy loss. Each dataset has a different number of classes: Cityscapes has 19 classes, Pascal VOC has 21 classes, ADE20K has 151 classes and COCO has 27 classes.

\paragraph{Depth Estimation.}
For monocular depth estimation, we train linear regressors on NYUv2 \citep{silberman2012nyuv2} for 20 epochs, with a learning rate of $5\times 10^{-4}$ and a one-cycle cosine annealing scheduler following depth estimation protocol of \citet{jafar}.
Consistent with the segmentation task, input and target images are resized to $448\times448$.
Ground-truth depth values are clipped to the range $[d_{\min}, d_{\max}]$, with $d_{\min} = 10^{-3}$ and $d_{\max} = 10.0$.

We optimize the model using a combination of scale-invariant and gradient-based losses.
Let $\hat{D}$ denote the predicted depth map and $D$ the target depth map, where values $D > d_{\max}$ are set to zero.
The total depth loss is defined as:
\begin{equation}
\mathcal{L}_{\text{depth}} = \lambda_{\sigma} \, \mathcal{L}_{\text{SI}}(\hat{D}, D) + \lambda_{\nabla} \, \mathcal{L}_{\text{grad}}(\hat{D}, D),
\end{equation}
where $\mathcal{L}_{\text{SI}}$ is the scale-invariant loss and $\mathcal{L}_{\text{grad}}$ is the gradient loss.
We set the weighting terms to $\lambda_{\sigma} = 10.0$ and $\lambda_{\nabla} = 0.5$ to encourage both accurate depth prediction and spatial smoothness.

\paragraph{Video Propagation.}
To evaluate the upsamplers in the context of video propagation, we follow the protocol of \citet{lift}. We use $448\times448$ input images and extracted features are extracted and subsequently upscaled by a factor of 2 using the proposed upsamplers, followed by bilinear interpolation.
Then we propagate labels using \citet{lift} protocol. Regarding the sparsity constraints, we set $k=20$, retaining only the 20 strongest source pixels for each target pixel based on affinity scores; all lower affinity values are zeroed, and we apply a binary locality constraint with a neighborhood radius of $r=24$.
This restricts the potential source pixels to a spatial window of size $(2r+1)^2$ centered around the target pixel before the top-$k$ selection is applied.

\paragraph{Open Vocabulary.}

We follow ProxyCLIP \citep{lan2024proxyclip} protocol using a $\times 4$ upsampling factor with $448 \times 448$ input images. Since it is direct inference, we did not change anything from the pipeline.

\subsection{Visualizations}

\begin{figure*}[ht!]
    \centering
    \includegraphics[width=\linewidth]{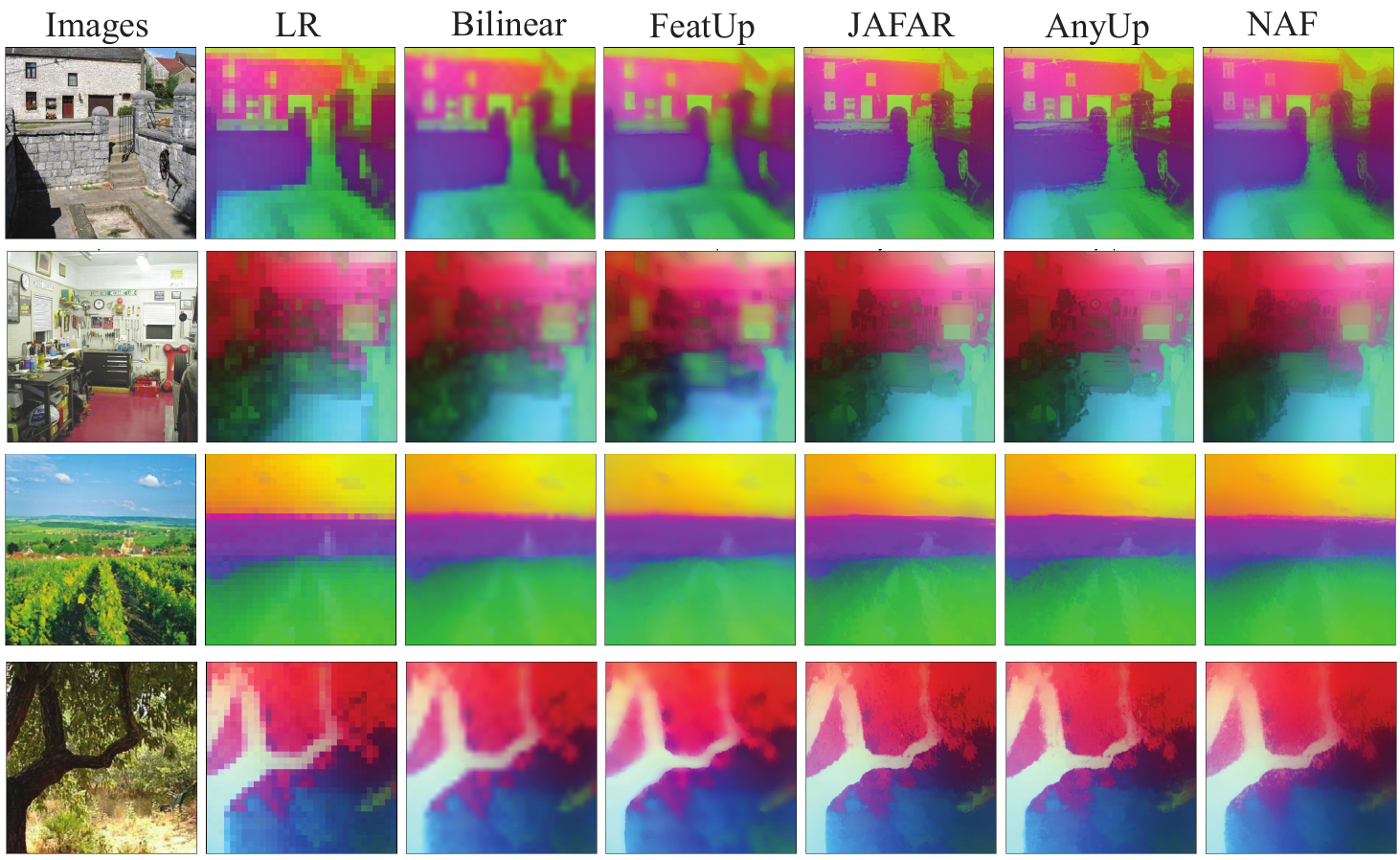}
    \caption{\textbf{PCA plots of different upsamplers for random images.} The first colum represents RGB images, the second one, the low resolution features, the others the PCA after upsampling. We use the same basis decomposition for plotting. Only JAFAR \cite{jafar}, AnyUp \cite{wimmer2025anyup} and \method{} produce sharp PCAs while preserving input feature representation.}
    \label{fig:pca}
\end{figure*}

\begin{figure*}[ht!]
    \centering
    \includegraphics[width=\linewidth]{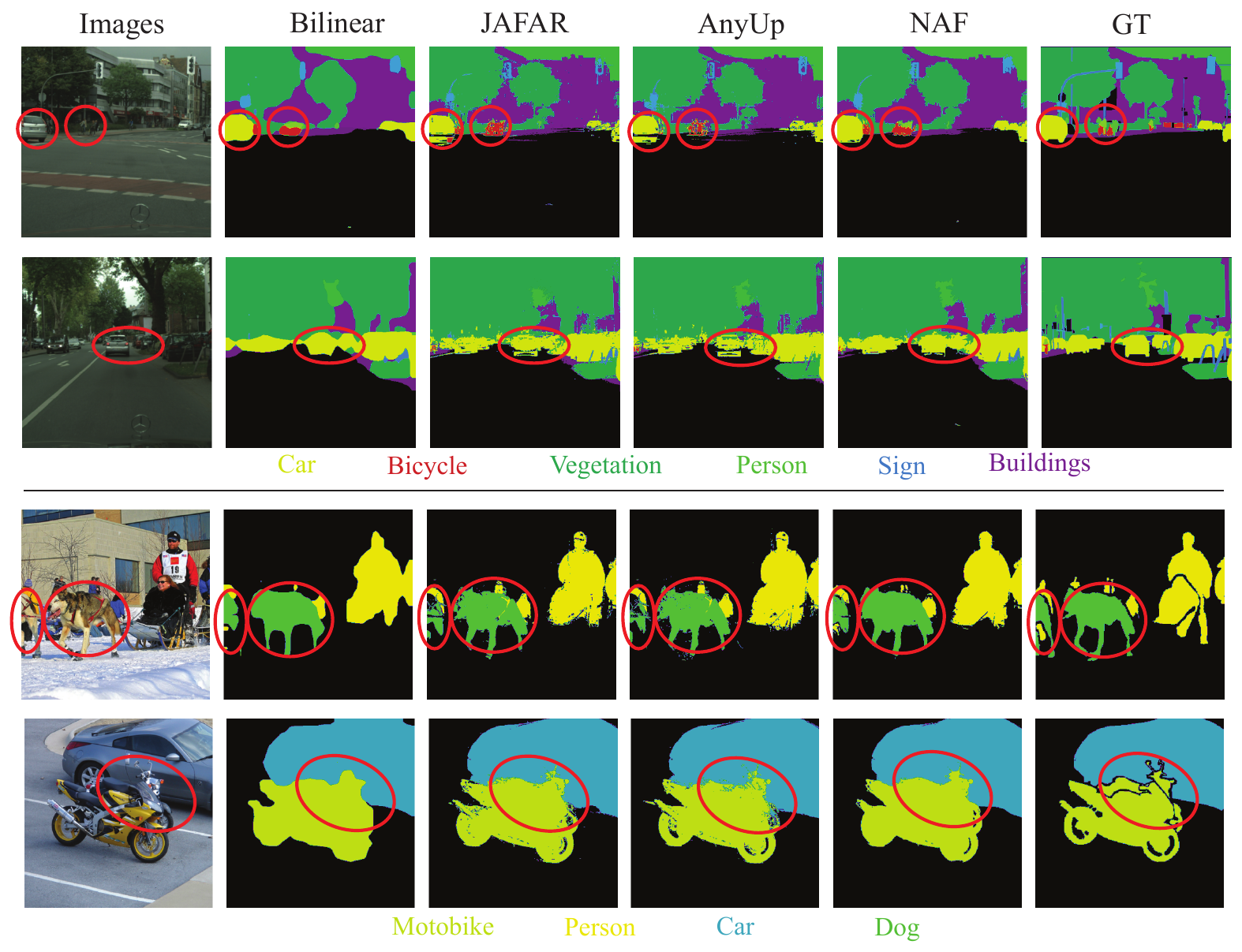}
    \caption{\textbf{Segmentation predictions using different upsamplers.} The first two rows are on Cityscapes \citep{cordts2016cityscapes}, the last two rows on VOC \citep{pascalvoc}. We see that while being VFM-agnostic \method{} better preserves structure compared to JAFAR \citep{jafar} and AnyUp \citep{wimmer2025anyup} that tend to focus too much on colors leading to noisy semantics.}
    \label{fig:segmentation}
\end{figure*}

\begin{figure*}[ht!]
    \centering
    \includegraphics[width=\linewidth]{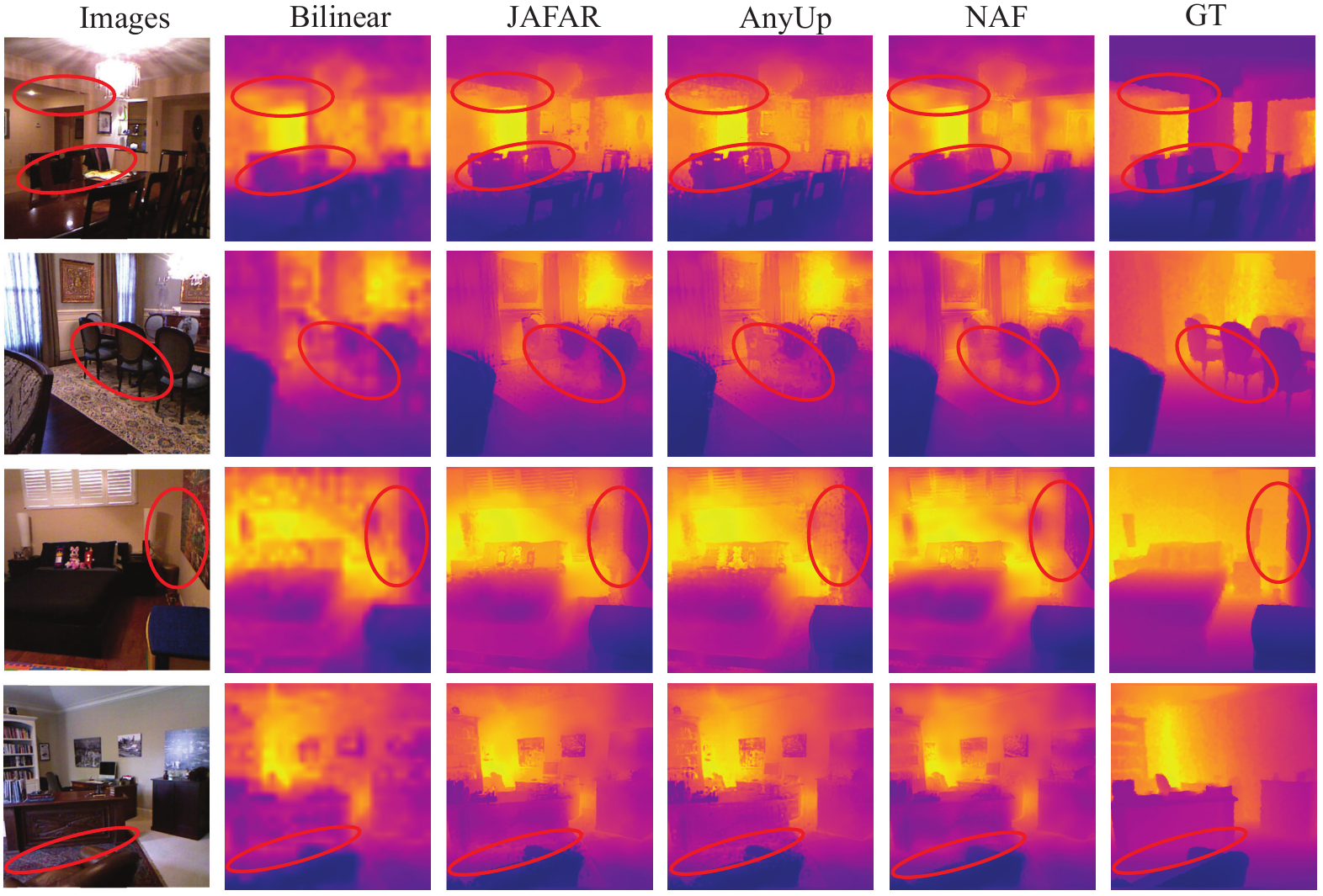}
    \caption{\textbf{Detph Estimation using different upsamplers} on NYUv2 \citep{silberman2012nyuv2}. Compared to AnyUp \cite{wimmer2025anyup}, \method{} outputs smoother predictions without the noisy effect we observe on some regions using AnyUp.}
    \label{fig:depth}
\end{figure*}

In \autoref{fig:pca}, we present PCA visualizations of the upsampled feature maps produced by the evaluated methods. The feature maps generated by \method{} exhibit smoother spatial variations compared to those of JAFAR \cite{jafar} and AnyUp \cite{wimmer2025anyup}, which display sharper transitions, while also avoiding the halo artifacts observed in FeatUp. These smoother feature maps are reflected in the downstream linear probing results for both semantic segmentation (\autoref{fig:segmentation}) and depth estimation (\autoref{fig:depth}). For segmentation, \method{} produces masks with substantially fewer sparse or fragmented artifacts compared to JAFAR and AnyUp. Likewise, the depth maps obtained with \method{} exhibit markedly smoother predictions in flat regions while still preserving sharp object boundaries, outperforming Bilinear, JAFAR and AnyUp in this regard.

\subsection{Generalization results}
\label{sec:supp_generalization}

To better assess the quality of upsamplers across different state-of-the-art VFMs, we evaluate a wide range of models of various sizes (T–S–B–L) from different families (PE-Core \citep{bolya2025PerceptionEncoder}, CAPI \citep{darcet2025capi}, DINO \citep{caron2021dino}, PE-Spatial \citep{bolya2025PerceptionEncoder}, SigLIP2 \citep{tschannen2025siglip}), using both VFM-specific and VFM-agnostic upsamplers.
on various datasets.
{To mitigate the computational cost of a full factorial evaluation, we adopt a randomized sampling strategy: two distinct VFMs are \emph{randomly} assigned to each dataset. The resulting performance metrics are summarized in \autoref{tab:generalization}.}
From this analysis, we draw the following conclusions:
\begin{itemize}
    \item Increasing the VFM input image size by a $\times 2$ factor leads to slightly higher average gains than using bilinear on standard images: \gain{+4.65 mIoU} vs \gain{+4.13 mIoU}. 
    \item The larger the scaling factor, the lower the results. Using $\times 4$ factor instead of $\times 2$ leads to lower results from \gain{+4.65 mIoU} gain to \loss{-6.25 mIoU} drop. Only DINOv3-L \citep{simeoni2025dinov3} on COCO continue to have higher results when increasing image size highlighting that increasing image size can increase scores depending on models and datasets. \method{} leads the average gains by \bestgain{+7.46 mIoU} resulting in a +1.23 mIoU gain over next previous state of the art.
\end{itemize}

\begin{table*}[ht!]
\centering
\resizebox{\textwidth}{!}{%
\begin{tabular}{l c cc cc cc cc c}
\toprule
 & \multicolumn{1}{c}{}&\multicolumn{2}{c}{\textbf{COCO} \citep{lin2014coco}} & \multicolumn{2}{c}{\textbf{VOC} \citep{pascalvoc}} & \multicolumn{2}{c}{\textbf{ADE20K}} \citep{fhou2017ade20k} & \multicolumn{2}{c}{\textbf{Cityscapes}} \citep{cordts2016cityscapes} & \multirow{2}{*}{$\Delta$ \textbf{Mean}} \\
\cmidrule(rl){3-4} \cmidrule(rl){5-6} \cmidrule(rl){7-8} \cmidrule(rl){9-10} 
\textbf{Method} & \textbf{V.A} & \textbf{DINOv3-L} & \textbf{DINOv2-R-S} & \textbf{CAPI-L} & \textbf{PE-Core-S} & \textbf{DINOv2-B} & \textbf{PE-Spatial-T} & \textbf{SigLIP2-B} & \textbf{DINOv3-C-S} \\
\midrule
Large ($\times 4$) & \cmark & \best{67.59} & 56.21 & 24.06 & 35.06 & 39.93 & 20.74 & 50.29 & \best{69.21} & \loss{-6.25} \\
\rowcolor{gray!25} LiFT \citep{lift} & \xmark & 61.45 & 57.66 & 78.82 & 55.36 & 38.75 & 17.31 & 52.02 & 48.02 & \loss{-0.46} \\
Nearest & \cmark & 64.47 & 57.34 & 76.99 & 59.25 & 39.82 & 22.24 & 45.05 & 47.90 & 0.00 \\
Bicubic \citep{keys2003bicubic} & \cmark & 66.55 & 58.98 & 79.75 & 62.76 & 41.93 & 23.35 & 49.61 & 54.70 & \gain{+3.07} \\
Bilinear & \cmark & 66.61 & 59.77 & 81.37 & 66.20 & 42.85 & 23.89 & 51.29 & 54.14 & \gain{+4.13} \\
Large ($\times 2$) & \cmark & \sbest{67.48} & 60.20 & 70.87 & 62.99 & 43.58 & \sbest{25.29} & \best{56.98} & \sbest{63.58} & \gain{+4.65} \\
\rowcolor{gray!25} FeatUp \citep{featup} & \xmark & 66.93 & 60.91 & 82.20 & 68.10 & 44.28 & 24.55 & 51.73 & 51.57 & \gain{+4.65}\\
\rowcolor{gray!25} FeatSharp \citep{featsharp}& \xmark & 66.85 & 59.80 & \sbest{84.10} & 67.35 & 42.90 & 23.61 & 52.75 & 56.84 & \gain{+5.14} \\
\rowcolor{gray!25} JAFAR \citep{jafar} & \xmark & 66.81 & \sbest{61.92} & 84.03 & \best{74.85} & \sbest{44.94} & 23.85 & 52.25 & 50.70 & \gain{+5.79} \\
AnyUp \citep{wimmer2025anyup}& \cmark & 66.54 & 61.58 & 84.00 & 73.66 & 44.44 & 24.68 & \sbest{53.58} & 54.40 & \gain{+6.23} \\
\rowcolor{lightpastelblue} {\method{}} & \cmark & {67.35} & \best{62.17} & \best{84.64} & \sbest{74.47} & \best{45.39} & \best{25.08} & \best{56.98} & 56.69 & \bestgain{+7.46} \\
\bottomrule
\end{tabular}%
}
\caption{\textbf{Semantic Segmentation (mIoU $\uparrow$) on Random Combinations of VFMs and datasets.} VFMs come from different sizes and families and are evaluated on many datasets: COCO \cite{lin2014coco}, Pascal VOC \citep{pascalvoc}, ADE20K \citep{fhou2017ade20k}. `\textbf{$\Delta$ Mean}' is computed against Nearest. We highlight \best{best} and \sbest{second best} scores, and \bestgain{best gain}. \textbf{V.A} indicates VFM-agnostic models.}
\label{tab:generalization}
\end{table*}

\subsection{Learning representation}

We investigate the following question for our framework: how does the choice of the VFM used during training affect the final upsampling performance? Although $\text{NAF}$ is designed for zero-shot use with any VFM at inference, its Dual-Branch Encoder is optimized using features from one specific VFM during training. To analyze this, we trained $\text{NAF}$ using features from DINOv3-B \citep{simeoni2025dinov3}, DINOv2-R-B \citep{darcet2023vitneedreg}, and DINOv2-S \citep{oquab2023dinov2}. The evaluation on other VFMs at inference time is detailed in \autoref{tab:ablation_learning}.
We find a counter-intuitive result: a VFM with strong general performance such as DINOv3-B does not necessarily yield the best results for training $\text{NAF}$, and we often achieve higher upsampling scores when training with smaller or less abstract representations such as DINOv2-R-S. Conversely, using raw RGB pixels (treating image upsampling as feature upsampling) caused a significant performance drop (`RGB'), confirming the need for encoded features. Furthermore, training with multiple VFMs simultaneously (`Mixture') did not improve scores (see. DINOv2-R-B), indicating that a single, appropriate representation is sufficient to efficiently guide the attention-based upsampling process.

\begin{table}[ht]
\centering
\resizebox{\columnwidth}{!}{%
\begin{tabular}{@{}l cccc c@{}}
\toprule
 \textbf{Backbone} & \textbf{DINOv2-R-B} & \textbf{RADIOv2.5-B} & \textbf{Franca-B} & \textbf{DINOv3-B} & \textbf{Mean} \\
\midrule
$\text{RGB}$ & 58.36 & 54.79 & 54.52 & 60.40 & 57.02 \\
No training & 60.46 & 56.53 & 56.26 & 61.92 & 58.79 \\
 \rowcolor{lightpastelblue} DINOv3-B & 63.82 & 58.76 & 58.03 & 61.01 & 60.41 \\
 DINOv2-S & 64.02 & 58.98 & 58.24 & 61.06 & 60.58 \\
Mixture & 64.16 & 59.18 & 58.29 & 61.73 & 60.84 \\
\rowcolor{Apricot!20!}  DINOv2-R-B & 65.25 & 59.86 & 59.24 & 62.54 & 61.72 \\
\bottomrule
\end{tabular}%
}
\caption{\textbf{Segmentation (mIoU $\uparrow$) on Cityscapes \citep{cordts2016cityscapes}}, training \method{} with different backbones. The best average score is highlighted in \textcolor{Apricot}{orange}, and the standard training configuration used for \method{} is indicated in \textcolor{lightpastelblue}{blue}.}
\label{tab:ablation_learning}
\end{table}

\subsection{VFM upsamplers as filters}

Instead of performing upsampling, we investigate how well learned upsamplers can function as feature filters. To do so, we apply the upsamplers using the target output resolution equal to that of the input features. In this setup, no upsampling is performed; the upsamplers effectively act as feature filters. The filtered features are subsequently upsampled using bilinear interpolation, and a linear classifier is trained on top of them. As shown in \autoref{tab:filters}, \method{} achieves the best average improvements compared to other filtering approaches with \bestgain{+0.75 mIoU}. Although the gains are modest, they suggest a “free lunch”: applying lightweight filters on top of VFMs yields additional mIoU without modifying the underlying model.

\begin{table}[ht!]
\centering
\resizebox{\columnwidth}{!}{%
\begin{tabular}{@{} l c c c c c c @{}}
\toprule
\textbf{Method} & \textbf{V.A} & \textbf{DINOv2-R} & \textbf{RADIO} & \textbf{Franca} & \textbf{DINOv3} & $\Delta$ \textbf{Mean} \\
\midrule
\rowcolor{gray!25} JAFAR \citep{jafar} & \xmark & 84.47 & \sbest{84.76} & 81.63 & 86.26 & \sbestgain{+0.32} \\
AnyUp \citep{wimmer2025anyup} & \cmark & \sbest{84.10} & 84.59 & \sbest{81.69} & 86.62 & \gain{+0.29} \\
Bilinear & \cmark & 83.07 & {84.47} &  81.30 &  \best{86.99}  & 0.00 \\
\rowcolor{lightpastelblue}
\method{} (ours) & \cmark & \best{84.90} & \best{85.05} & \best{82.01} & \sbest{86.88} & \bestgain{+0.75} \\
\bottomrule
\end{tabular}%
}
\caption{\textbf{Semantic Segmentation (mIoU $\uparrow$) on VOC \citep{pascalvoc} using VFM-upsamplers as feature filters.} We use base VFMs: DINOv2-R-B \citep{darcet2023vitneedreg}, RADIOv2.5-B \citep{heinrich2025radiov25}, Franca-B \citep{venkataramanan2025franca}, and DINOv3-B \citep{simeoni2025dinov3}.
\textbf{$\Delta$ Mean} is computed relative to Bilinear. \best{Bold} and \sbest{underline} indicate the best and second-best scores, respectively, while \bestgain{highlighted} indicates the largest gain. \textbf{V.A} denotes VFM-agnostic models.}
\label{tab:filters}
\end{table}

\subsection{Scaling ratio}

We evaluate the robustness of \method{} to different upsampling ratio in \autoref{tab:seg_scale_probing}. We take as input different image resolution ($224\times224$, $448\times448$ and $896\times896$), feed them to the VFM and upscale the features to $448\times448$ resolution before learning the linear probing classifier.

\method{} always leads to the best or second best score regardless of the scaling ratio, proving that it can be used across a wide range of resolutions. Interestingly, as already mentioned, for some VFM (e.g., PE-Core-B) feeding larger images does not lead to higher scores. Nonetheless, \method{} still improves the mIoU of degraded representations.
\begin{table}[t]
\centering
\resizebox{\columnwidth}{!}{%
\begin{tabular}{@{} l c c c | c c c @{}}
\toprule
\multirow{2}{*}{\textbf{Method}} & \multicolumn{3}{c|}{\textbf{Radiov2.5-B}} & \multicolumn{3}{c}{\textbf{PE-Core-B}} \\
 & 14 $\rightarrow$ 448 & 28 $\rightarrow$ 448 & 56 $\rightarrow$ 448 & 14 $\rightarrow$ 448 & 28 $\rightarrow$ 448 & 56 $\rightarrow$ 448 \\
\midrule
AnyUp & 53.85 & 62.04 & \oomcell & \best{54.92} & \sbest{58.64} & \oomcell \\
Bilinear & \sbest{55.85} & \sbest{66.63} & \sbest{68.74} & 53.54 & 56.83 & 43.81 \\
Nearest & 49.10 & 61.34 & 66.42 & 45.12 & 49.98 & 39.03 \\
\rowcolor{gray!25} JAFAR & 54.51 & 62.33 & \oomcell & 51.02 & 51.95 & \oomcell \\
\rowcolor{Apricot!20!} \method{} & \best{56.31} & \best{67.02} & \best{69.04} & \sbest{54.47} & \best{61.06} & \best{48.44} \\
\bottomrule
\end{tabular}%
}
\caption{\textbf{Semantic Segmentation (mIoU $\uparrow$) on Cityscapes \citep{cordts2016cityscapes} for different Upsampling-ratio}. We compare different upsamplers for generating $448 \times 448$ features from various feature input resolutions. The first three columns correspond to RADIOv2.5-B \citep{heinrich2025radiov25}, and the last three columns to PE-Core-B \citep{cho2025PerceptionLM}. \best{Bold} indicates the best score, and \sbest{underline} the second-best. \oomcell{} indicates linear probing-training `Out-of-Memory'.}
\label{tab:seg_scale_probing}
\end{table}

\section{Image Restoration}

\subsection{Evaluation Setup}
\label{sec:supp_restoration}

We evaluate several image denoising models --- DNCNN \citep{zhang2017dncnn}, IRCNN \citep{zhang2017ircnn}, RedNet \citep{jiang2018rednet}, and Restormer \citep{Zamir2021Restormer} --- on ImageNet for $25$k steps using input images of size $448 \!\times\! 448$. We select the largest batch size that fits on a single A100 40GB GPU: $B=32$ for the convolutional models and $B=1$ for Restormer \citep{Zamir2021Restormer}.

The denoisers are trained with a combination of three loss terms: L1, L2, and SSIM. Denoting the total loss as
\begin{equation}
\mathcal{L} = \lambda_1 \, \mathcal{L}_{\text{L1}} + \lambda_2 \, \mathcal{L}_{\text{L2}} + \lambda_3 \, \mathcal{L}_{\text{SSIM}},
\end{equation}
we set the weights to $\lambda_1 = 1.0$, $\lambda_2 = 5.0$, and $\lambda_3 = 0.2$. To train \method{} we keep the same architecture than for feature upsampling but we increase the neighborhood kernel size from 9 to 15 to take into account the receptive field difference.

\subsection{Visualizations}

\begin{figure}[ht!]
    \centering
    \includegraphics[width=\columnwidth]{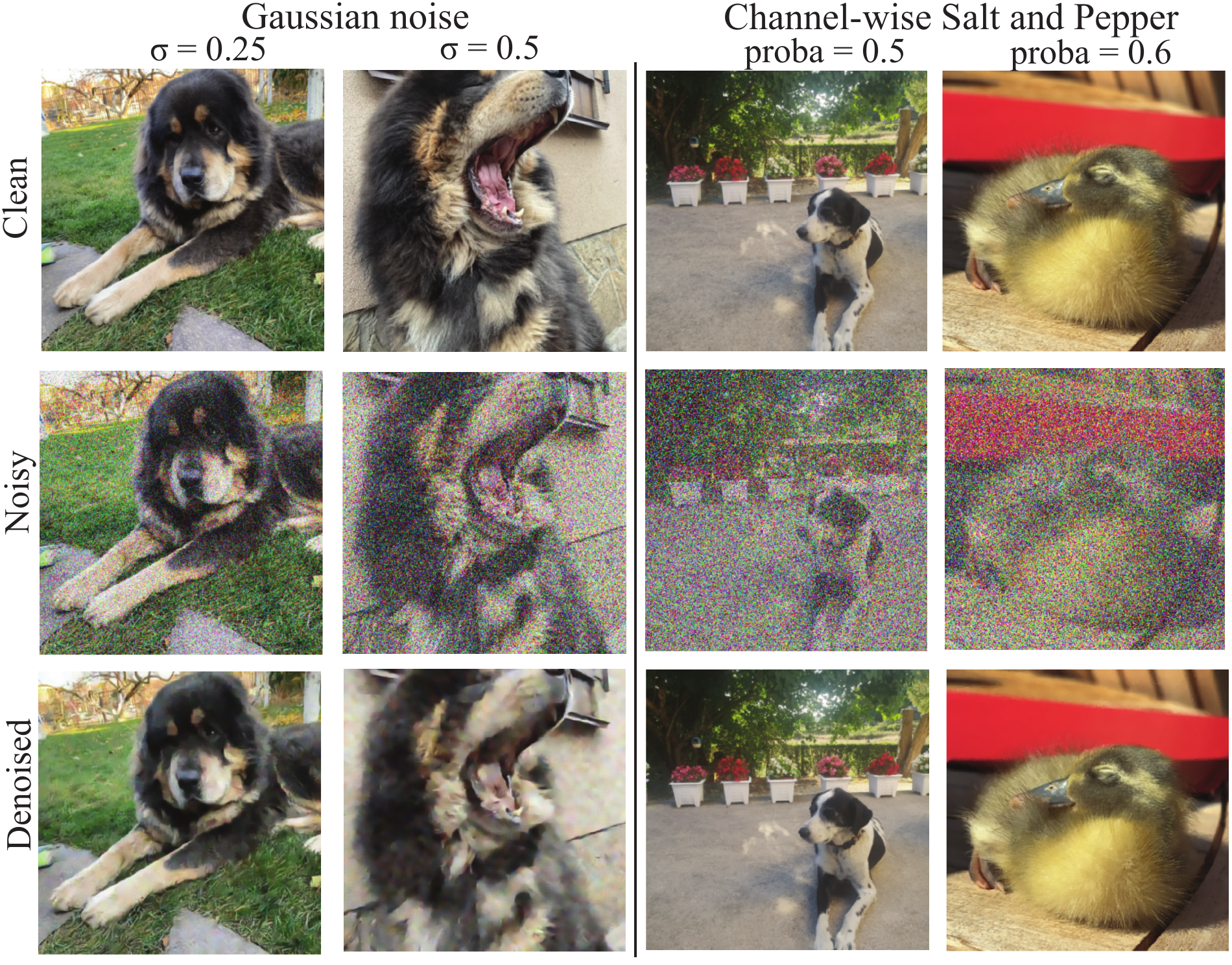}
    \caption{\textbf{Image restoration using \method{}}. On the left two images we apply a gaussian noise. On the right we apply a channel-wise salt and pepper noise. \method{} allows to restore very noisy images even on unseen noise range (rightmost image).}
    \label{fig:denoising}
\end{figure}

To evaluate the denoiser’s performance, we apply noise to a set of clean $448 \times 448$ images and feed them to \method{}. In \autoref{fig:denoising}, we test models trained on dynamic ranges of Gaussian noise $\sigma \in [0.1, 0.5]$ and salt-and-pepper noise $p \in [0.1, 0.5]$. We observe that the model can effectively denoise channel-wise salt-and-pepper noise even beyond the maximal training range (rightmost image uses 0.6, while the model has been trained up to 0.5 noise intensity), while achieving high-quality reconstructions for other noise levels as well.

\section{Computation footprint}

\begin{figure*}[ht!]
    \centering
    \begin{subfigure}[b]{0.32\linewidth}
        \centering
        \includegraphics[width=\linewidth]{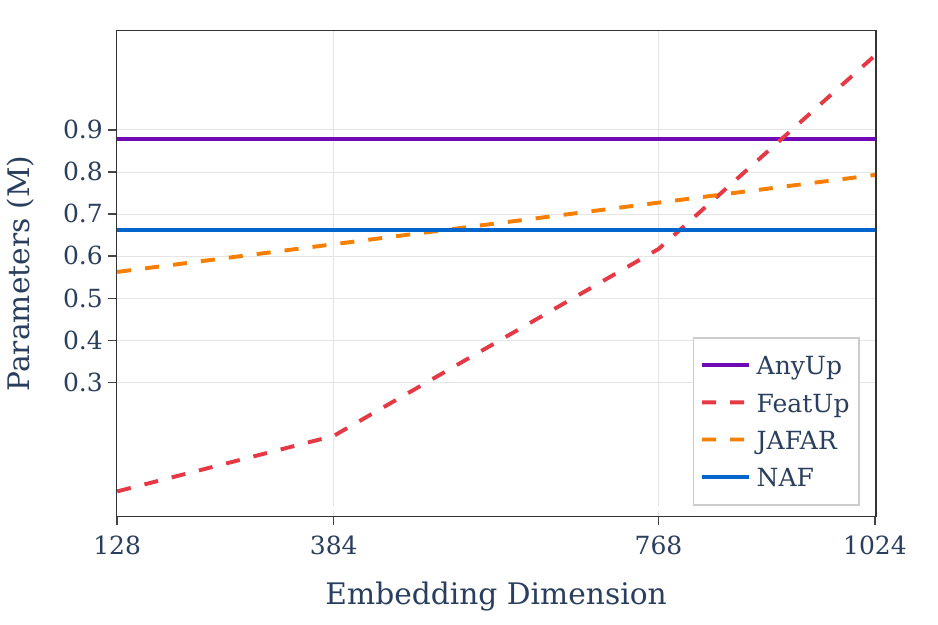}
        \caption{Number of Parameters}
    \end{subfigure}
    \hfill
    \begin{subfigure}[b]{0.32\linewidth}
        \centering
        \includegraphics[width=\linewidth]{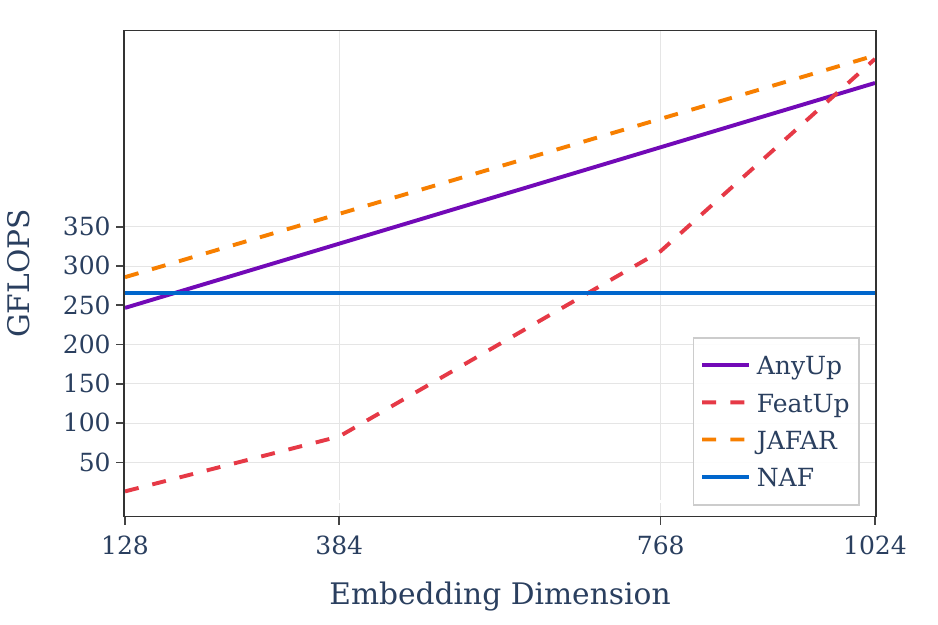}
        \caption{Computational Cost (GFLOPs)}
    \end{subfigure}
    \hfill
    \begin{subfigure}[b]{0.32\linewidth}
        \centering
        \includegraphics[width=\linewidth]{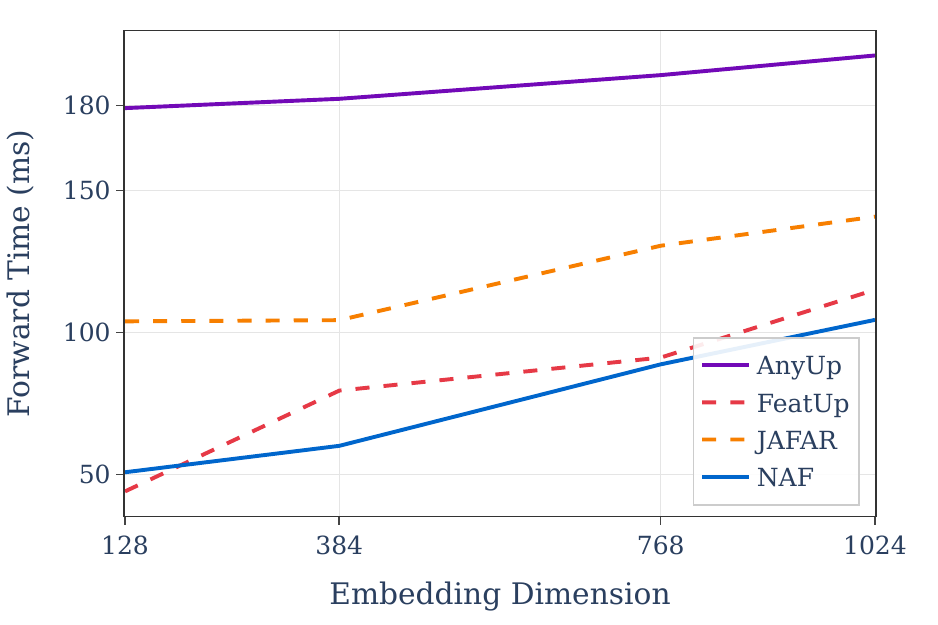}
        \caption{Avg. Forward Pass Time}
    \end{subfigure}

    \vspace{1em} %

    \begin{subfigure}[b]{0.32\linewidth}
        \centering
        \includegraphics[width=\linewidth]{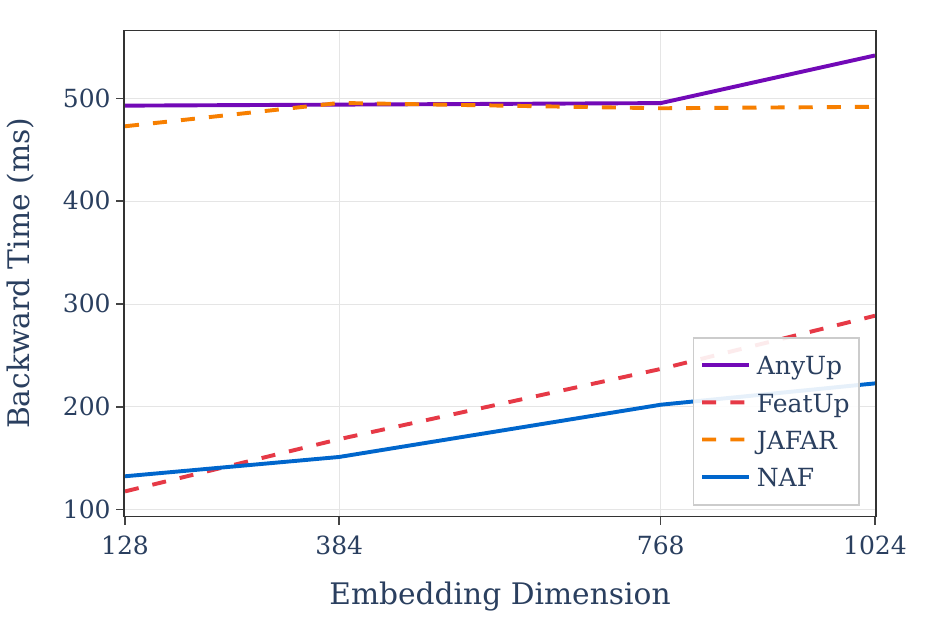}
        \caption{Avg. Backward Pass Time}
    \end{subfigure}
    \hfill
    \begin{subfigure}[b]{0.32\linewidth}
        \centering
        \includegraphics[width=\linewidth]{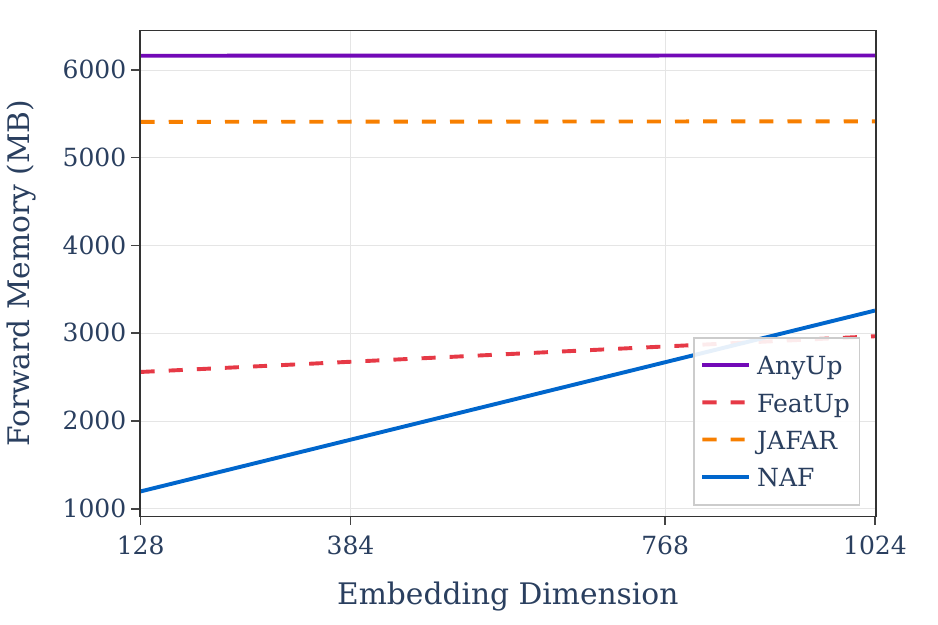}
        \caption{Peak Memory (Forward)}
    \end{subfigure}
    \hfill
    \begin{subfigure}[b]{0.32\linewidth}
        \centering
        \includegraphics[width=\linewidth]{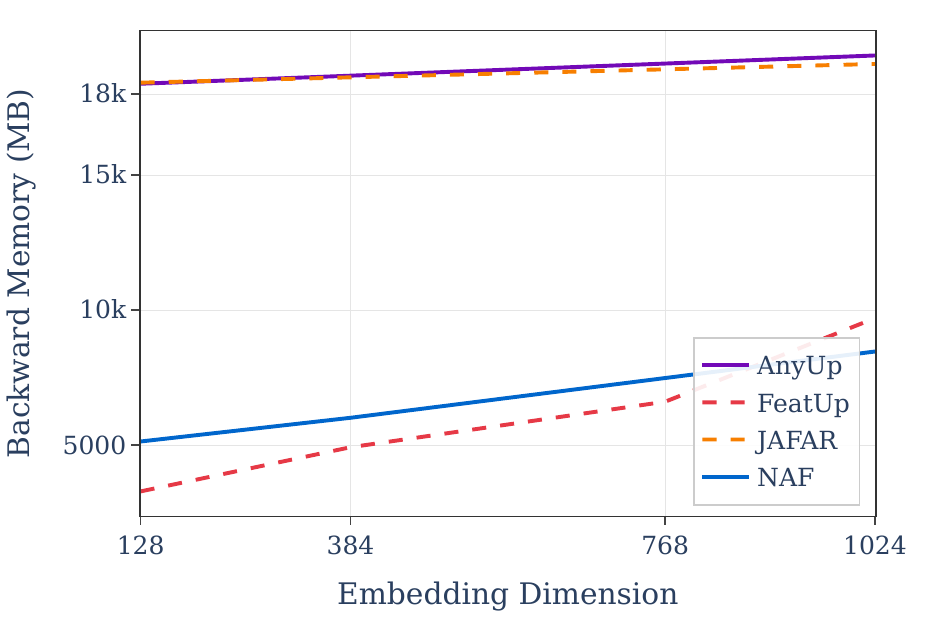}
        \caption{Peak Memory (Backward)}
    \end{subfigure}

    \caption{\textbf{Benchmarking analysis across embedding dimensions.} We study 4 different standard embedding sizes: 128, 384, 768 and 1024. (a)-(b) show model complexity, (c)-(d) compare execution time, and (e)-(f) analyze memory consumption.}
    \label{fig:embed_results}
\end{figure*}

\begin{figure*}[ht!]
    \centering
    \begin{subfigure}[b]{0.32\linewidth}
        \centering
        \includegraphics[width=\linewidth]{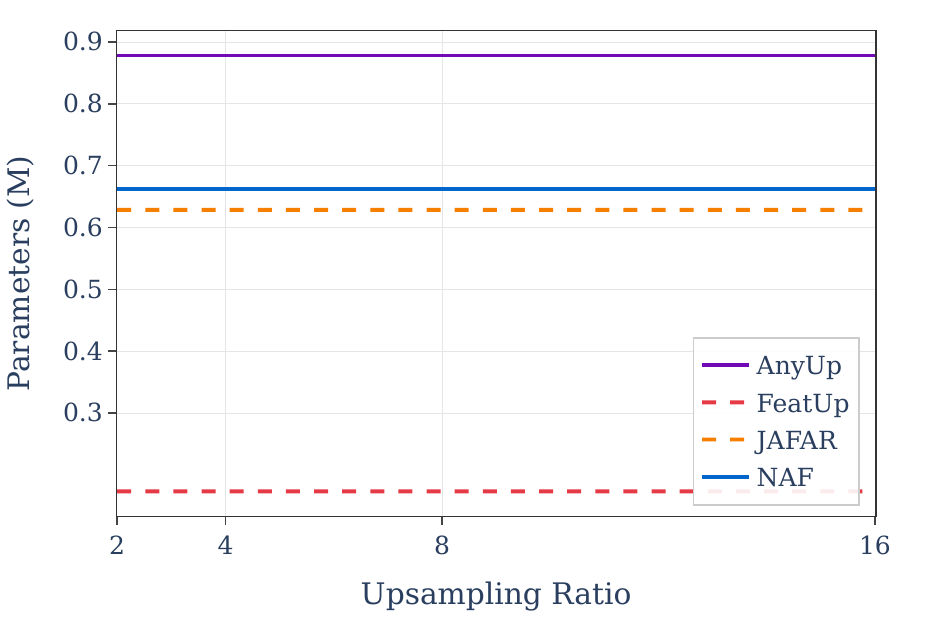}
        \caption{Number of Parameters}
    \end{subfigure}
    \hfill
    \begin{subfigure}[b]{0.32\linewidth}
        \centering
        \includegraphics[width=\linewidth]{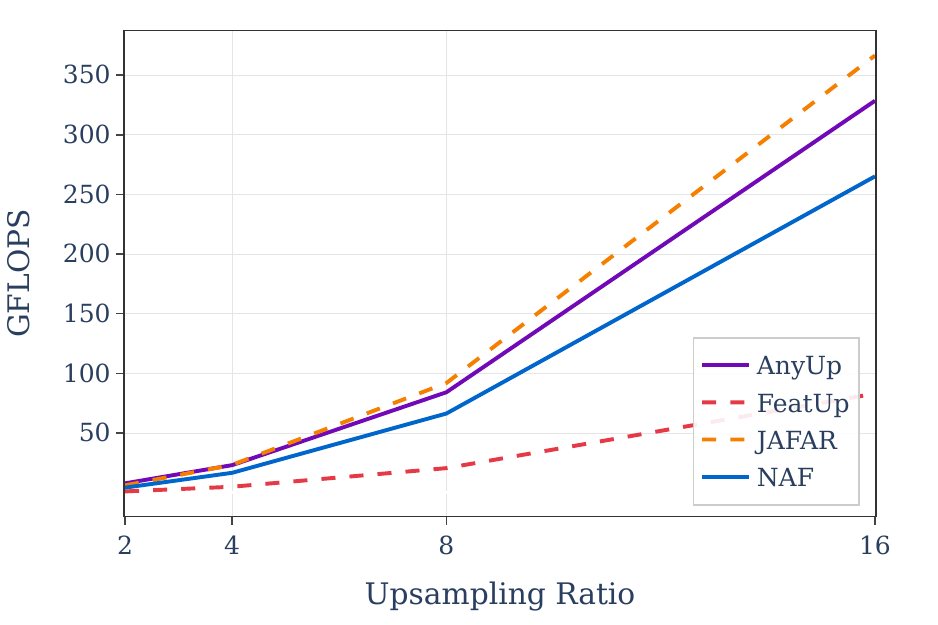}
        \caption{Computational Cost (GFLOPs)}
    \end{subfigure}
    \hfill
    \begin{subfigure}[b]{0.32\linewidth}
        \centering
        \includegraphics[width=\linewidth]{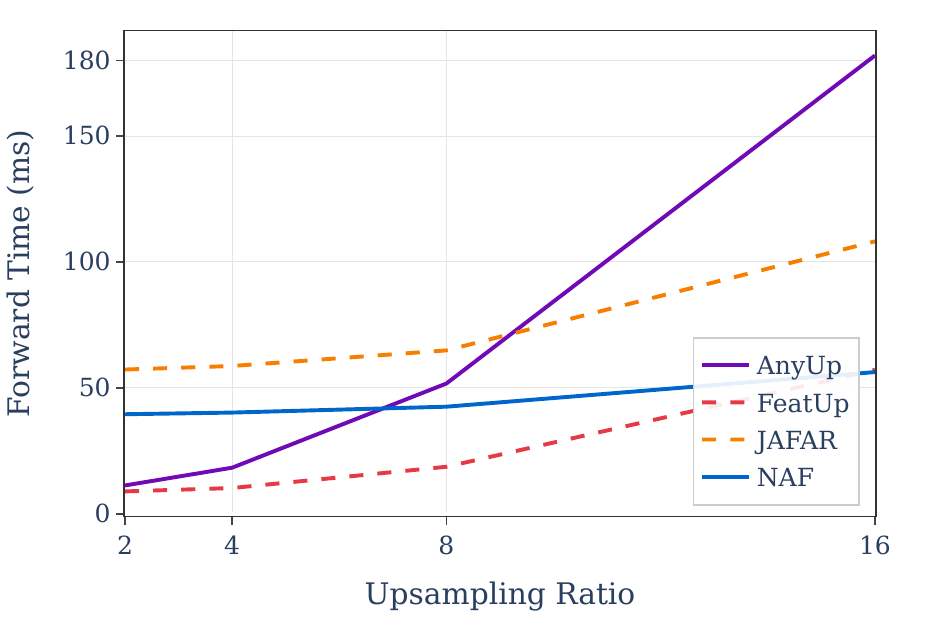}
        \caption{Avg. Forward Pass Time}
    \end{subfigure}

    \vspace{1em} %

    \begin{subfigure}[b]{0.32\linewidth}
        \centering
        \includegraphics[width=\linewidth]{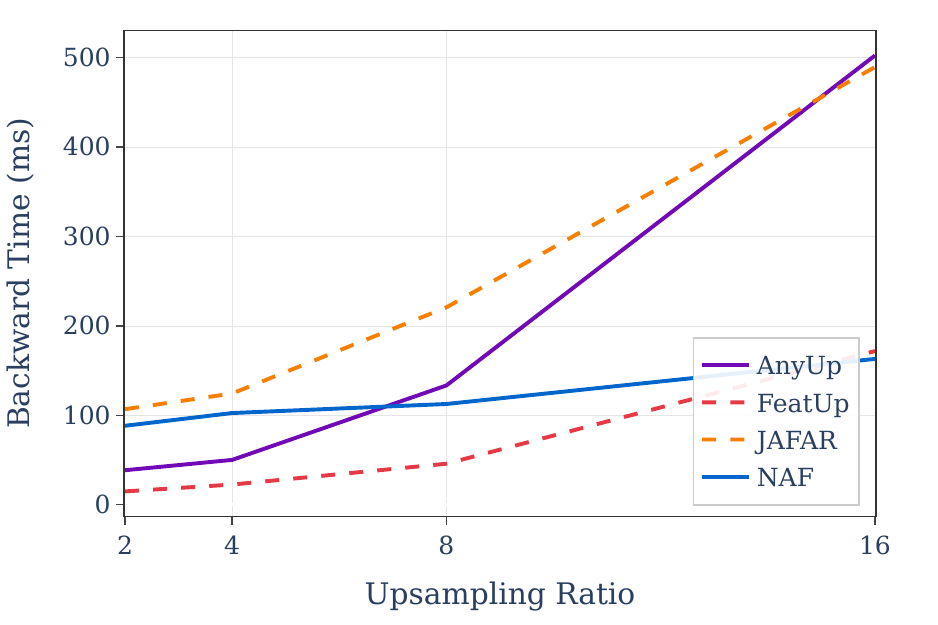}
        \caption{Avg. Backward Pass Time}
    \end{subfigure}
    \hfill
    \begin{subfigure}[b]{0.32\linewidth}
        \centering
        \includegraphics[width=\linewidth]{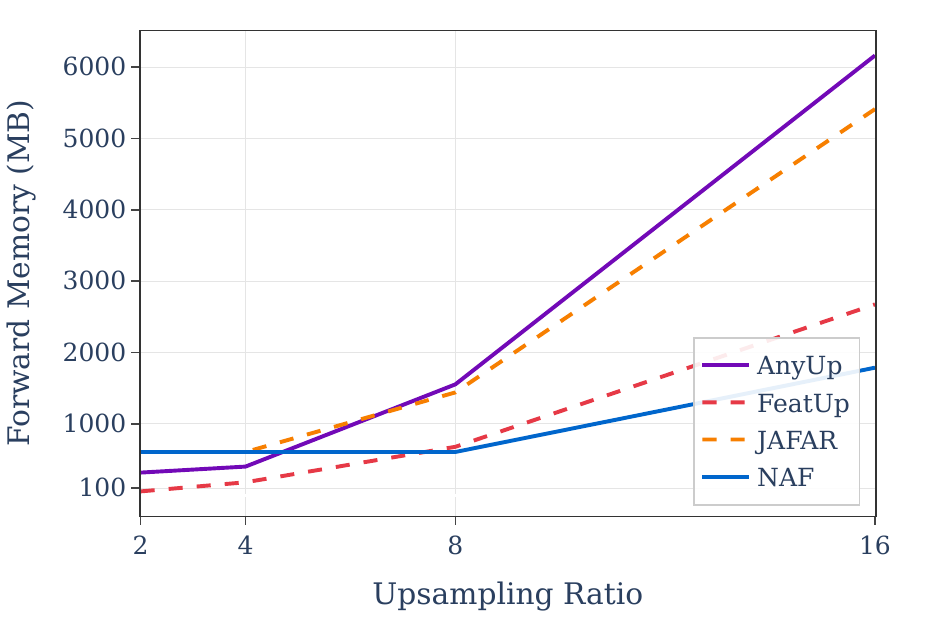}
        \caption{Peak Memory (Forward)}
    \end{subfigure}
    \hfill
    \begin{subfigure}[b]{0.32\linewidth}
        \centering
        \includegraphics[width=\linewidth]{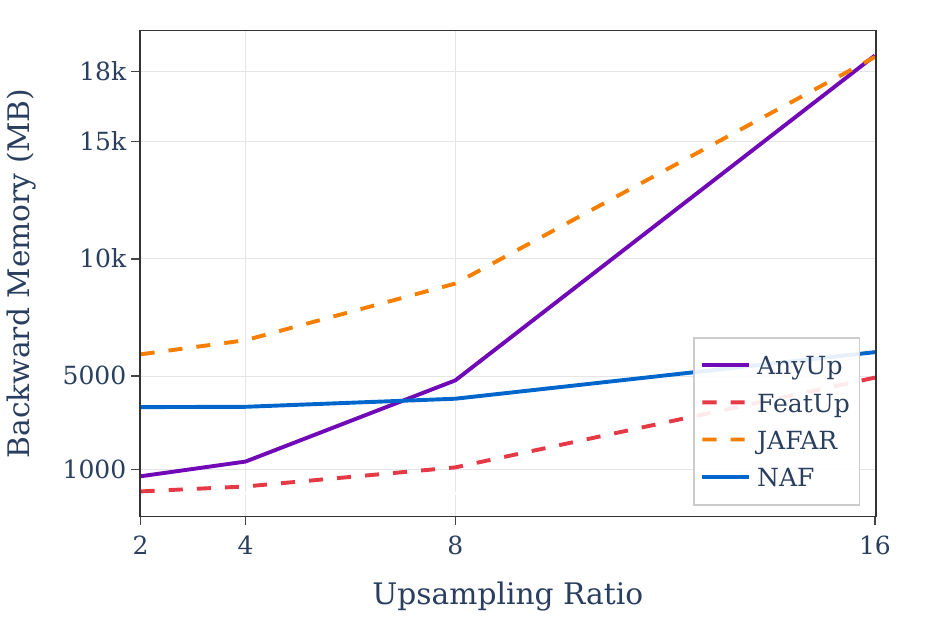}
        \caption{Peak Memory (Backward)}
    \end{subfigure}

    \caption{\textbf{Benchmarking analysis across upscaling ratio.} We studied different upscaling ratio: $\times 2$, $\times 4$, $\times 8$, $\times 16$. (a)-(b) show model complexity, (c)-(d) compare execution time, and (e)-(f) analyze memory consumption.}
    \label{fig:ratio_results}
\end{figure*}

We previously provided initial insights into the efficiency of different upsamplers in \autoref{tab:caracteristics}, where \method{} is the only approach capable of achieving a $\times 72$ upsampling ratio, producing features at a resolution of $2048 \times 2048$.
We now investigate in greater detail the behavior of these upsamplers when targeting different scaling factors or when processing higher-dimensional feature maps (\autoref{fig:embed_results}, \autoref{fig:ratio_results}). 
To this end, we initialize a dummy feature tensor of size $(28, 28, 384)$, corresponding to the output of a standard model processing $448 \times 448$ input images. We then conduct two controlled studies: (i) varying the embedding dimension while keeping a fixed $\times 16$ upsampling ratio (\autoref{fig:embed_results}), and (ii) varying the upsampling ratio while maintaining 384 channels (\autoref{fig:ratio_results}). For each configuration, we evaluate the total number of parameters (in millions, M), the computational cost (GFLOPs), the time required for the forward pass (relevant for inference) and backward pass (relevant for training), as well as the peak memory consumption during both forward and backward passes.

Although JAFAR \citep{jafar} and AnyUp \citep{wimmer2025anyup} are attention-based models like \method{}, \method{} achieves substantially higher memory and computational efficiency. In the embedding study, it provides an approximately $2$–$3\times$ speedup and memory reduction compared to AnyUp. For low upsampling ratios, AnyUp is slightly more efficient; however, its efficiency degrades rapidly with larger upsampling factors, resulting in an approximately $3\times$ advantage for \method{} at high upscaling levels. 
Furthermore, processing larger input images with a standard state-of-the-art VFM leads to a substantial increase in GFLOPs and runtime, whereas \method{} maintains significantly lower computational cost, as shown in \autoref{tab:naf_vs_large}.

\begin{table}[t]
\centering
\resizebox{0.7\columnwidth}{!}{%
\begin{tabular}{@{}lccc ccc@{}}
\toprule
\multirow{2}{*}{\textbf{Method}} & \multicolumn{3}{c}{\textbf{GFLOPs}} & \multicolumn{3}{c}{\textbf{Time (ms)}} \\
\cmidrule(lr){2-4} \cmidrule(lr){5-7}
 & ×2 & ×4 & ×8 & ×2 & ×4 & ×8 \\
\midrule
\textbf{Large} & 537 & 2148 & 8591 & 110 & 1035 & 6555 \\
\rowcolor{lightpastelblue} \textbf{NAF} & 4 & 16 & 66 & 39 & 40 & 42 \\
\bottomrule
\end{tabular}
}
\caption{\textbf{Comparison of NAF and the Large Image baseline.}
We measure GFLOPs and inference time required to produce features at $\times 2$, $\times 4$, and $\times 8$ the original resolution of DINOv3-B \citep{simeoni2025dinov3} outputs. For the Large Image baseline, this corresponds to feeding images scaled by the same factors, relative to the standard $448 \times 448$ resolution.}
\label{tab:naf_vs_large}
\end{table}

Across many configurations, the efficiency of \method{} is comparable to that of FeatUp \citep{featup}, which relies on convolutions but is intrinsically constrained to fixed output resolutions. In contrast, \method{} combines the flexibility of an any-scale upsampler with the computational efficiency characteristic of convolution-based designs, while consistently outperforming attention-based alternatives.

\section{Limitations and Perspectives}

Compared to both VFM-specific and VFM-agnostic upsamplers, \method{} achieves state-of-the-art performance across multiple datasets and tasks. Nevertheless, several avenues remain for improvement.

By design, \method{} employs a neighborhood-attention mechanism with a fixed kernel size. We observe that the attention maps vary depending on the query point (see. \autoref{fig:attention_maps}). Introducing dynamic kernel adaptation—similar to approaches in Deformable Attention or Deformable Convolutions—could, in principle, reduce the computational cost per interpolation step while potentially enhancing reconstruction accuracy by introducing sampling flexibility.

Although our method is VFM-agnostic, we currently lack a principled framework for identifying which VFMs provide the most informative patch representations for learning the upsampling. Closing this gap requires a deeper understanding of the representational properties that support zero-shot consistent upsampling. Empirically, we observe that neither combining multiple VFMs nor using larger or stronger VFMs leads to clear gains  (see. \autoref{tab:ablation_learning}).

In terms of applications, looking ahead, the ability to preserve high-resolution spatial representations is especially valuable in medical imaging and remote sensing, highlighting promising avenues for future research. In addition, we have demonstrated that \method{}'s architecture is versatile and can be adapted across domains, particularly within the denoising context, paving the way to other applications such as in image restoration.

\end{document}